%% file: main.tex
\documentclass{article}

\usepackage{microtype}
\usepackage{graphicx}
\usepackage{subcaption}
\usepackage{booktabs} 
\usepackage{multirow}
\usepackage{titlesec}

\usepackage{hyperref}
\usepackage{amsmath}
\DeclareMathOperator*{\argmax}{arg\,max}

\usepackage[accepted]{icml2024}

\usepackage{amsmath}
\usepackage{amssymb}
\usepackage{mathtools}
\usepackage{amsthm}
\usepackage{algorithm}
\usepackage{algorithmic}

\usepackage[capitalize,noabbrev]{cleveref}

\theoremstyle{plain}

\theoremstyle{definition}

\theoremstyle{remark}

\usepackage[textsize=tiny]{todonotes}

\newcommand{\rebuttalrevision}{}

\icmltitlerunning{Actions Speak Louder than Words: Trillion-Parameter Sequential Transducers for Generative Recommendations}

\begin{document}

\twocolumn[
\icmltitle{Actions Speak Louder than Words: Trillion-Parameter Sequential Transducers for Generative Recommendations}




\begin{icmlauthorlist}
\icmlauthor{Jiaqi Zhai}{meta}
\icmlauthor{Lucy Liao}{meta}
\icmlauthor{Xing Liu}{meta}
\icmlauthor{Yueming Wang}{meta}
\icmlauthor{Rui Li}{meta} \\
\icmlauthor{Xuan Cao}{meta}
\icmlauthor{Leon Gao}{meta}
\icmlauthor{Zhaojie Gong}{meta}
\icmlauthor{Fangda Gu}{meta}
\icmlauthor{Michael He}{meta} 
\icmlauthor{Yinghai Lu}{meta} 
\icmlauthor{Yu Shi}{meta}
\end{icmlauthorlist}

\icmlaffiliation{meta}{MRS, Meta AI} %

\icmlcorrespondingauthor{}{\{jiaqiz, lucyyl, xingl, yuemingw, ruili\}@meta.com}

\icmlkeywords{Generative Recommenders, Generative Recommendations, Sequential Recommendations, Generative Modeling, Actions Speak Louder Than Words, HSTU, Hierarchical Sequential Transduction Units, Recommendation Systems, Recommendations, Machine Learning, ICML}

\vskip 0.3in
]



\printAffiliationsAndNotice{}  

\input{sec_abstract}

\input{sec_intro}
\input{sec_design}
\input{sec_evaluations_v2}
\input{sec_related_work}
\input{sec_conclusions}
\input{sec_acknowledgements}

\bibliography{example_paper}
\bibliographystyle{icml2024}

\newpage
\appendix
\onecolumn
\input{sec_appendix}

\end{document}

%% file: sec_abstract.tex

\begin{abstract}
Large-scale recommendation systems are characterized by their reliance on high cardinality, heterogeneous features and the need to handle tens of billions of user actions on a daily basis. Despite being trained on huge volume of data with thousands of features,
most Deep Learning Recommendation Models (DLRMs) in industry fail to scale with compute.
Inspired by success achieved by Transformers in language and vision domains, we revisit fundamental design choices in recommendation systems. We reformulate recommendation problems as sequential transduction tasks
within a generative modeling framework (``Generative Recommenders''), and propose a new architecture, HSTU, designed for 
high cardinality, non-stationary streaming recommendation data.
HSTU outperforms baselines over synthetic and public datasets by up to 65.8\% in NDCG, \rebuttalrevision{and is 5.3x to 15.2x faster than FlashAttention2-based Transformers on 8192 length sequences.}
HSTU-based Generative Recommenders, with
1.5 trillion parameters, improve metrics in online A/B tests by 12.4\% and have been deployed on multiple surfaces of a large internet platform with billions of users. \rebuttalrevision{More importantly,
the model quality of Generative Recommenders empirically scales as a power-law of training compute across three orders of magnitude,
up to GPT-3/LLaMa-2 scale,
which reduces carbon footprint needed for future model developments, and further paves the way for the first foundation models in recommendations.}
\vspace{-1em}
\end{abstract}

%% file: sec_intro.tex
\section{Introduction}

\begin{figure}[h]
    \vspace{-.9em} 
    \begin{center}
    \includegraphics[width=0.78\linewidth]{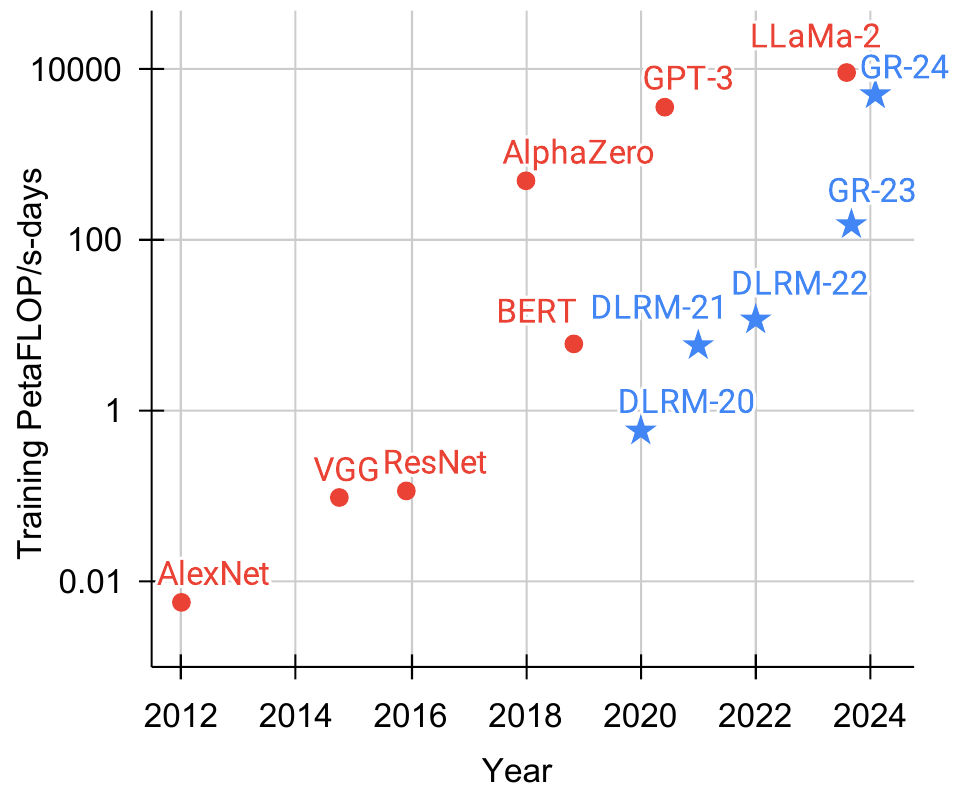}
    \end{center}
    \vspace{-1.7em}
    \caption{Total compute used to train deep learning models over the years. DLRM results are from~\citep{dlrm_isca22}; GRs are deployed models from this work. DLRMs/GRs are continuously trained in a streaming setting; we report compute used per year.}
    \label{fig:intro_dl_models_pf_days}
    \vspace{-1.6em}
\end{figure}

Recommendation systems, quintessential in the realm of online content platforms and e-commerce, play a pivotal role in personalizing billions of user experiences on a daily basis. 
State-of-the-art approaches in recommendations have been based on Deep Learning Recommendation Models (DLRMs)~\citep{dlrm_isca22} for about a decade~\citep{ytdnn_goog_recsys16,wdl_goog_dlrs16,din_baba_kdd18,ple_recsys20,dcnv2_www21,pinterest23transact}. 
DLRMs are characterized by their usage of heterogeneous features, such as numerical features -- counters and ratios, embeddings, and categorical features such as creator ids, user ids, etc. Due to new content and products being added every minute, the feature space
is of extreme high cardinality, often in the range of billions~\citep{pixie_pins_www18}. To leverage tens of thousands of such features, DLRMs employ various neural networks to combine features, transform intermediate representations, and \rebuttalrevision{compose the final outputs}.

Despite utilizing extensive human-engineered feature sets and training on vast amounts of data, 
most DLRMs in industry scale poorly with compute~\citep{zhao2023ucranking}. This limitation is noteworthy and remains unanswered.
 

\rebuttalrevision{Inspired by the success achieved by Transformers in language and vision, we revisit fundamental design choices in modern recommendation systems.} We observe that
alternative formulations at billion-user scale need to overcome three challenges. First, features in recommendation systems lack explicit structures. While sequential formulations have been explored in small-scale settings \rebuttalrevision{(detailed discussions in~\cref{sec:app-generative-recommenders-background-and-formulations})},
heterogeneous features, including high cardinality ids, cross features, counters, ratios, etc. play critical roles in industry-scale DLRMs~\citep{dlrm_isca22}. Second, recommendation systems use billion-scale vocabularies that change continuously. A billion-scale dynamic vocabulary, in contrast to 100K-scale static ones in language~\citep{brown2020language}, creates training challenges and necessitates high inference cost given the need to consider tens of thousands of candidates in a target-aware fashion~\citep{din_baba_kdd18,wang2020cold_preranking}. Finally, computational cost represents the main bottleneck in enabling large-scale sequential models. GPT-3 was trained on a total of 300B tokens over a period of 1-2 months with thousands of GPUs~\citep{brown2020language}. This scale appears daunting, until we contrast it with the scale of user actions. The largest internet platforms serve billions of daily active users, who engage with billions of posts, images, and videos per day. User sequences could be of length up to 
$10^5$~\citep{kwai2023twin}.  
Consequentially, recommendation systems need to handle a few orders of magnitude more tokens \textit{per day} than what language models process over 1-2 months.

In this work, we treat \textit{user actions} as a new \textit{modality} in generative modeling. Our key insights are, a) core ranking and retrieval tasks in industrial-scale recommenders can be cast as generative modeling problems given an appropriate new feature space; b) this paradigm enables us to systematically leverage redundancies in features, training, and inference to improve efficiency. 
Due to our new formulation, we deployed models that are \textit{three orders of magnitude} more computationally complex than prior state-of-the-art, while improving topline metrics by 12.4\%, as shown in~\cref{fig:intro_dl_models_pf_days}.


Our contributions are as follows. We first propose \textit{Generative Recommenders} (GRs) in~\cref{sec:dlrms-to-grs}, a new paradigm replacing DLRMs. We sequentialize and unify the heterogeneous feature space in DLRMs, with the new approach approximating the full DLRM feature space as sequence length tends to infinity. This enables us to reformulate the main recommendation problems, ranking and retrieval, as pure sequential transduction tasks in GRs. Importantly, this further enables model training to be done in a sequential, generative fashion, which permits us to train on orders of magnitude more data \textit{with the same amount of compute}. 

We next address computational cost challenges throughout training and inference. We propose a new sequential transduction architecture, \textit{Hierarchical Sequential Transduction Units} (HSTU). HSTU modifies attention mechanism for large, non-stationary vocabulary, and exploits characteristics of recommendation datasets to achieve 5.3x to \rebuttalrevision{15.2x speedup vs FlashAttention2-based Transformers on 8192 length sequences}. Further, through a new algorithm, M-FALCON, that fully amortizes computational costs via micro-batching (\cref{sec:design_e2e_inference}), \rebuttalrevision{we can serve \textit{285x more complex} GR models while achieving 1.50x-2.99x speedups, all \textit{with the same inference budget} used by traditional DLRMs.}

We finally validate the proposed techniques over synthetic datasets, public datasets, and deployments on multiple surfaces of a large internet platform with billions of daily active users in~\cref{sec:experiments}. 
To the best of our knowledge, our work represents the first result that shows pure sequential transduction-based architectures, like HSTU, in generative settings (GRs) to significantly outperform DLRMs in large-scale
industrial settings. 
Remarkably, not only did we overcome known scaling bottlenecks
in traditional DLRMs, we further succeeded in showing that scaling law~\citep{kaplan2020scalinglaws} applies to recommendations, representing the potential ChatGPT moment for recommendation systems.

%% file: sec_design.tex

\vspace{-.5em}
\section{Recommendation as Sequential Transduction Tasks: From DLRMs to GRs} 
\label{sec:dlrms-to-grs}
\subsection{Unifying heterogeneous feature spaces in DLRMs}
\label{sec:dlrms-to-grs-features}

\begin{figure*}[t!]
  \centering
  \vspace{-.8em}
  \includegraphics[width=0.68\linewidth]{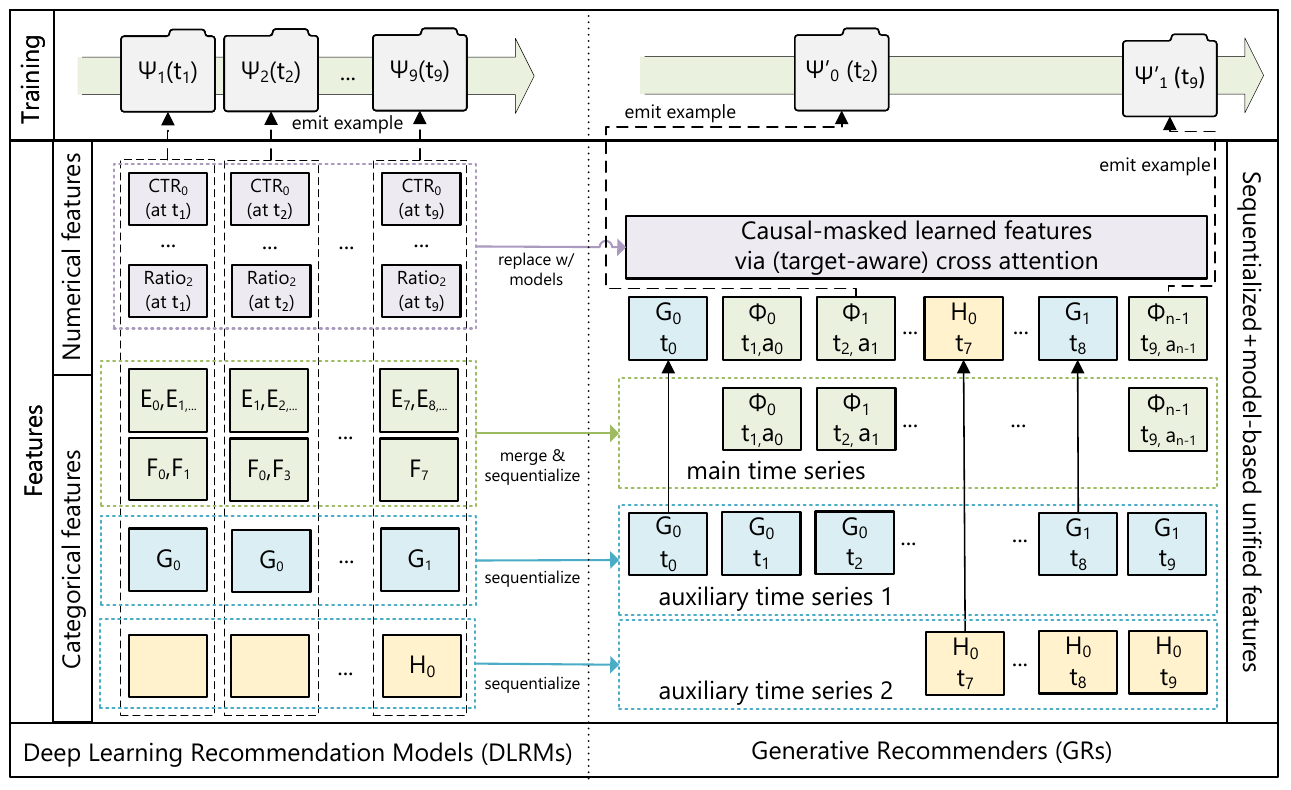} 
  \vspace{-1.2em}
  \caption{Comparison of features and training procedures: DLRMs vs GRs.
  $E,F,G,H$ denote categorical features. $\Phi_i$ represents the $i$-th item in the merged main time series. $\Psi_k(t_j)$ denotes training example $k$ emitted at time \rebuttalrevision{$t_j$. Full notations can be found in~\cref{sec:app-notations}.}}
  \label{fig:design_dlrms_grs_features_training}
   \vspace{-1.2em}
\end{figure*}


Modern DLRM models are usually trained with a vast number of categorical (`sparse') and numerical (`dense') features.
In GRs, we consolidate and encode these features into a single unified time series, as depicted in~\cref{fig:design_dlrms_grs_features_training}. 

\textbf{Categorical (`sparse') features}. Examples of such features
include items that user liked, creators in a category (e.g., Outdoors)
that user is following, user languages, communities that user joined, cities from which requests were initiated, etc. We \textit{sequentialize} these features as follows. We first select the longest time series, typically by merging the features that represent items user engaged with, as the main time series. The remaining features are generally time series that slowly change over time, such as demographics or followed creators.
We \textit{compress} these time series by keeping the earliest entry per consecutive segment and then merge the results into the main time series. Given these time series change very slowly, this approach does not significantly increase the overall sequence length.

\textbf{Numerical (`dense') features}. Examples of such features include weighted and decayed counters, ratios, etc. For instance, one feature could represent user's past click through rate (CTR) on items matching a given topic. Compared to categorical features, these features change much more frequently, potentially with every single (user, item) interaction. It is therefore infeasible to fully sequentialize such features from computational and storage perspectives. However,
an important observation is that
the categorical features (e.g., item topics, locations) over which we perform these aggregations are already sequentialized and encoded in GRs.
Hence, we can remove numerical features in GRs given a sufficiently expressive sequential transduction architecture \textit{coupled with a target-aware formulation}~\citep{din_baba_kdd18} can meaningfully capture numerical features
as we increase the overall sequence length and compute in GRs.


\vspace{-.3em}
\subsection{Reformulating ranking and retrieval as sequential transduction tasks}
\label{sec:dlrms-to-grs-tasks}


\rebuttalrevision{Given a list of $n$ tokens $x_0, x_1, \ldots, x_{n-1}$ ($x_i \in \mathbb{X}$) ordered chronologically, the time when those tokens are observed $t_0, t_1, \ldots, t_{n-1}$, a sequential transduction task maps this input sequence to the output tokens $y_0, y_1, \ldots, y_{n-1}$ ($y_i \in \mathbb{X} \cup \{ \varnothing \}$), where $y_i = \varnothing$ indicates that $y_i$ is undefined.}

\rebuttalrevision{We use $\Phi_i \in \mathbb{X}_c$ ($\mathbb{X}_c \subseteq \mathbb{X}$) to denote a content (e.g., images or videos) that the system provides to the user. 
Given new content are constantly created, $\mathbb{X}_c$ and $\mathbb{X}$ are non-stationary. The user can respond to $\Phi_i$ with some action $a_i$ (e.g., like, skip, video completion+share) $a_i \in \mathbb{X}$. We denote the total number of contents that a user has interacted with by $n_c$.}

\rebuttalrevision{The standard ranking and retrieval tasks, in causal autoregressive settings, can then be defined as sequential transduction tasks (\cref{tbl:dlrms-to-grs-task-definitions}). 
We make the following observations:}

\begin{table}[hb]
\small
\vspace{-0.8em}
\begin{center}
\begin{tabular}{lll}
\toprule
\multicolumn{1}{c}{\bf Task}  & \multicolumn{2}{c}{\bf Specification (Inputs / Outputs)} \\
\hline 
    \multirow{2}{*}{Ranking} & $x_i$s & $\Phi_0,a_0,\Phi_1,a_1,\ldots,\Phi_{n_c-1},a_{n_c-1}$   \\
                             & $y_i$s  & $a_0,\varnothing,a_1,\varnothing,\ldots,a_{n_c-1},\varnothing$\\
    \midrule
    \multirow{3}{*}{Retrieval} & $x_i$s & $(\Phi_0,a_0), (\Phi_1,a_1), \ldots, (\Phi_{n_c-1},a_{n_c-1})$ \\
                             & \multirow{2}{*}{$y_i$s} & $\Phi_1',\Phi_2',\ldots,\Phi_{n_c-1}',\varnothing$  \\
                             &              & ($\Phi_i' = \Phi_i$ if $a_i$ is positive, otherwise $\varnothing$) \\
\bottomrule
\end{tabular}
\vspace{-0.5em}
\caption{\rebuttalrevision{Ranking and retrieval as sequential transduction tasks. Other categorical features are omitted for simplicity. We compare GRs with traditional sequential recommenders in~\cref{sec:app-grs-formulations}.}}
\label{tbl:dlrms-to-grs-task-definitions}
\end{center}
\vspace{-1.5em}
\end{table}

\textbf{Retrieval.} In recommendation system's retrieval stage, we learn a distribution $p(\Phi_{i+1}|u_i)$ over $\Phi_{i+1} \in \mathbb{X}_c$, where $u_i$ is the user's representation \rebuttalrevision{at token $i$}. A typical objective is to select $\argmax_{\Phi \in \mathbb{X}_c} p(\Phi|u_i)$ to maximize some reward. This differs from a standard autoregressive setup in two ways. First, the supervision for $x_i$, $y_i$, is not necessarily $\Phi_{i+1}$, as users could respond negatively to $\Phi_{i+1}$. Second, $y_i$ is undefined when $x_{i+1}$ represents a non-engagement related categorical feature, such as demographics. 

\textbf{Ranking.} Ranking tasks in GRs pose unique challenges as industrial recommendation systems often require a ``target-aware'' formulation. In such settings, ``interaction'' of target, $\Phi_{i+1}$, and historical features 
needs to occur as early as possible, which is infeasible with a standard autoregressive setup where ``interaction'' happens late (e.g., via softmax after encoder output). We address this by \textit{interleaving} items and actions \rebuttalrevision{in~\cref{tbl:dlrms-to-grs-task-definitions}, which enables the ranking task to be formulated as $p(a_{i+1} | \Phi_0, a_0, \Phi_1, a_1, \ldots, \Phi_{i+1})$ (before categorical features)}. 
We apply a small neural network to transform outputs at $\Phi_{i+1}$ 
into multi-task predictions in practice. Importantly, this enables us to apply target-aware cross-attention to all $n_c$ engagements in one pass.


\vspace{-.3em}
\subsection{Generative training}
\label{sec:dlrms-to-grs-generative-training}


Industrial recommenders are commonly trained in a streaming setup, where each example 
is processed sequentially as they become available. 
In this setup, the total computational requirement for self-attention based sequential transduction architectures, such as Transformers~\citep{transformers_goog_neurips17}, scales as $
\sum_i n_i(n_i^2 d + n_i d_{ff} d)$, where $n_i$ is the number of tokens of user $i$, and $d$ is the embedding dimension.
The first part in the parentheses 
comes from self-attention, with assumed $O(n^2)$ scaling factor due to most subquadratic algorithms involving quality tradeoffs and underperforming quadratic algorithms in wall-clock time~\citep{dao2022flashattention}. The second part comes from pointwise MLP layers, with hidden layers of size $O(d_{ff})=O(d)$. Taking $N = \max_i n_i$, the overall time complexity reduces to $O(N^3d + N^2 d^2)$, which is cost prohibitive for recommendation settings.

To tackle the challenge of training sequential transduction models over long sequences in a scalable manner, we move from traditional impression-level training to \textit{generative training}, reducing the computational complexity by an $O(N)$ factor, as shown at the top of Figure~\ref{fig:design_dlrms_grs_features_training}. By doing so, encoder costs are amortized across multiple targets. More specifically, when we sample the $i$-th user at rate $s_u(n_i)$, the total training cost now scales as
$
\sum_i s_u(n_i) n_i (n_i^2d + n_i d^2)
$, which is reduced to $O(N^2d + N d^2)$ by setting $s_u(n_i)$ to $1/n_i$. One way to implement this sampling in industrial-scale systems is to emit training examples at the end of a user's request or session, resulting in $\hat{s_u}(n_i) \propto 1/n_i$.

\section{A High Performance Self-Attention Encoder for Generative Recommendations}
\label{sec:design_hstu_encoder}


To scale up GRs for industrial-scale recommendation systems with 
large, non-stationary vocabularies, 
we next introduce a new
encoder design, 
\textit{Hierarchical Sequential Transduction Unit} (HSTU).
HSTU consists of 
a stack of identical layers connected
by residual connections~\citep{He2015_resnet}.
Each layer contains three 
sub-layers: Pointwise Projection (Equation~\ref{eq:hstu-pointwise-projection}), Spatial Aggregation (Equation~\ref{eq:hstu-spatial-aggregation}), and Pointwise Transformation (Equation~\ref{eq:hstu-pointwise-transformation}):
\begin{equation}
\small
    U(X), V(X), Q(X), K(X) = \text{Split}(\phi_1(f_1(X)))
    \label{eq:hstu-pointwise-projection}
\end{equation}
\vspace{-1em}
\begin{equation}
\small
    A(X)V(X) = \phi_2\left(Q(X)K(X)^T + \text{rab}^{p,t}\right)V(X)
    \label{eq:hstu-spatial-aggregation}
\end{equation}
\vspace{-1em}
\begin{equation}
\small
    Y(X) = f_2\left(\text{Norm}\left(A(X)V(X)\right) \odot U(X)\right)
    \label{eq:hstu-pointwise-transformation}
\end{equation}

where $f_i(X)$ denotes an MLP; we use one linear layer, $f_i(X) = W_i(X) + b_i$ for $f_1$ and $f_2$ to reduce compute complexity \rebuttalrevision{and further batches computations for queries $Q(X)$, keys $K(X)$, values $V(X)$, and gating weights $U(X)$ with a fused kernel}; $\phi_1$ and $\phi_2$ denote nonlinearity, for both of which we use $\text{SiLU}$~\citep{elfwing2017silu}; \rebuttalrevision{$\text{Norm}$ is layer norm;} and $\text{rab}^{p,t}$ denotes relative attention bias~\citep{t5_jmlr20} that incorporates positional ($p$) and temporal ($t$) information.
\rebuttalrevision{Full notations used can be found in \cref{tbl:table-of-notations-continued}.} 

\begin{figure}[h]
    \vspace{-0.5em}
\begin{center}
  \includegraphics[width=0.9\linewidth]{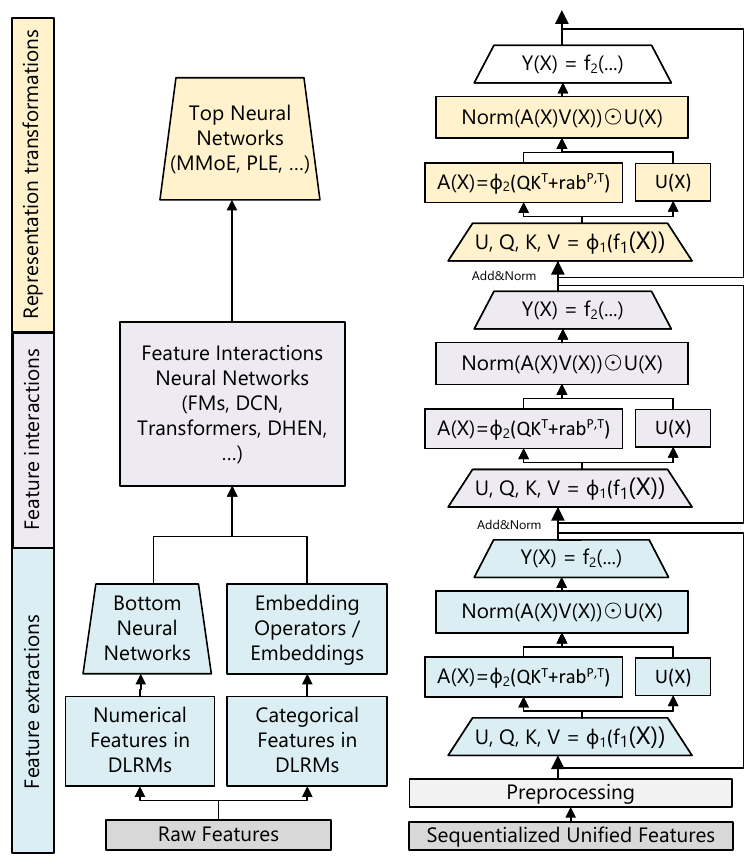}
\end{center}
    \vspace{-1em}
\caption{Comparison of key model components: DLRMs vs GRs. The complete DLRM setup~\citep{dlrm_isca22} is shown on the left side and a simplified HSTU is shown on the right.}
  \label{fig:dlrms_vs_grs_model_components}
  \vspace{-1.2em}
\end{figure}

 HSTU encoder design allows for the replacement of heterogeneous
 modules in DLRMs with a single modular block. We observe that there are, effectively, three main stages in
 DLRMs: \textit{Feature Extraction}, \textit{Feature Interactions}, and
 \textit{Transformations of Representations}. \textit{Feature Extractions} retrieves the pooled embedding representations of categorical features. Their most advanced versions can be generalized as pairwise attention and target-aware pooling~\citep{din_baba_kdd18}, which is captured with HSTU layers. 

\textit{Feature Interaction} is the most critical part of DLRMs. Common approaches used 
include factorization machines and their neural network variants
~\citep{fm_rendle_icdm10,deepfm_ijcai17,afm_ijcai17}, higher order feature interactions~\citep{dcnv2_www21}, etc. 
HSTU replaces feature interactions by enabling attention pooled features to directly ``interact'' with other features via $\text{Norm}\left(A(X)V(X)\right) \odot U(X)$. This design is motivated by the difficulty of approximating dot products with learned MLPs~\citep{ncf_mf_goog_recsys20,meta23ndp}. Given SiLU is applied to $U(X)$, $\text{Norm}\left(A(X)V(X)\right) \odot U(X)$ can also be interpreted as a variant of  SwiGLU~\citep{shazeer2020glu}.

\textit{Transformations of Representations} is commonly done with Mixture of Experts (MoEs) and routing to handle diverse, heterogeneous populations. A key idea 
is to perform conditional computations by specializing sub-networks for different users~\citep{mmoe_kdd18,ple_recsys20}. Element-wise dot products in HSTU can virtually perform gating operations used in MoEs up to a normalization factor.

\vspace{-.3em}
\subsection{Pointwise aggregated attention}
\label{sec:pointwise-silu-attention}

HSTU adopts a new pointwise aggregated (\rebuttalrevision{normalized}) attention mechanism (in contrast, softmax attention \rebuttalrevision{computes normalization factor over the entire sequence}). 
This is motivated by two factors. First, 
the number of prior data points related to target serves as a strong feature indicating the intensity of user preferences, which is hard to capture after softmax normalization. This is critical as we need to predict both the intensity of engagements, e.g., time spent on a given item, and the relative ordering of the items, e.g., predicting an ordering 
to maximize AUC. Second, while softmax activation is robust to noise by construction, it is less suited for non-stationary vocabularies in streaming settings.



The proposed pointwise aggregated attention mechanism is depicted in~\cref{eq:hstu-spatial-aggregation}.
Importantly, layer norm is needed after pointwise pooling to stabilize training.
One way to understand this design is through synthetic data following a Dirichlet Process that generates streaming data over a nonstationary vocabulary (details in Appendix~\ref{sec:app-exp-synthetic-data}). In this setting, we can observe gaps as large as 44.7\% between softmax and pointwise attention setups 
as shown in~\cref{tbl:synthetic-data}.

\begin{table}[t]
\small
\vspace{-.5em}
\begin{center}
\begin{tabular}{lll}
\toprule
\multicolumn{1}{c}{\bf Architecture}  &\multicolumn{1}{c}{\bf HR@10} &\multicolumn{1}{c}{\bf HR@50}
\\ \hline
        Transformers & .0442 & .2025 \\
        HSTU (-$\text{rab}^{p,t}$, Softmax) & .0617 & .2496 \\
        HSTU (-$\text{rab}^{p,t}$) & \textbf{.0893} & \textbf{.3170} \\
\bottomrule
\end{tabular}
\end{center}
\vspace{-1em}
\caption{Synthetic data in one-pass streaming settings.}
\label{tbl:synthetic-data}
\vspace{-1.2em}
\end{table}

\subsection{Leveraging and algorithmically increasing sparsity}
\label{sec:design-algorithmic-sparsity}
In recommendation systems, the length of user history sequences often follows a skewed distribution, leading to sparse input sequences, particularly in the settings with very long sequences. This sparsity can be leveraged to significantly improve the efficiency of the encoder.
To achieve this, we have developed an efficient attention kernel for GPUs that fuses back-to-back GEMMs in a manner similar to~\citep{rabe2021memoryefficientselfattention,dao2022flashattention}
but performs \textit{fully raggified} attention computations. This essentially transforms
the attention computation into grouped GEMMs of various sizes (\rebuttalrevision{\cref{sec:app-hstu-kernels}}).
As a result, self-attention in HSTU becomes memory-bound and scales as
$\Theta(\sum_i n_i^2 d_{qk}^2 R^{-1})$ in terms of memory accesses, where $n_i$ is
the sequence length for sample $i$, $d_{qk}$ is attention dimension, and $R$ is the register size. This approach 
by itself leads to 2-5x throughput gains as discussed in~\cref{sec:exp-encoder-efficiency}.


We further \textit{algorithmically} increase the sparsity of user history sequences via \textit{Stochastic Length} (SL). One key characteristic of user history sequences in recommendations is that user behaviors are temporally repetitive, 
as user behaviors manifest at multiple scales throughout their interaction history.
This represents an opportunity to increase sparsity \textit{artificially} without
compromising model quality, thereby significantly reducing encoder cost that scales as $\Theta(\sum_i n_i^2)$.

\rebuttalrevision{We can represent user $j$'s history as a sequence $(x_i)_{i=0}^{n_{c, j}}$, where $n_{c,j}$ is the number of contents user interacted with. Let $N_c = \max_j {n_{c,j}}$. 
Let $(x_{i_k})_{k=0}^{L}$ be a subsequence of length $L$ constructed from the original sequence $(x_i)_{i=0}^{n_{c,j}}$.} $SL$ selects input sequences as follows:
\vspace{-0.5em}
\begin{equation}
\small
\begin{split}
\rebuttalrevision{(x_i)_{i=0}^{n_{c,j}}} & \text{ if } n_{c,j} \leq N_c^{\alpha/2}  \\
\rebuttalrevision{(x_{i_k})_{k=0}^{N_c^{\alpha/2}}} & \text{ if } n_{c,j} > N_c^{\alpha/2}, \text{w/ probability } 1 - N_c^\alpha / n_{c,j}^2 \\
\rebuttalrevision{(x_i)_{i=0}^{n_{c,j}}} & \text{ if } n_{c,j} > N_c^{\alpha/2}, \text{w/ probability } N_c^\alpha / n_{c,j}^2
\end{split}
\label{eq:sl_sampling}
\end{equation}
which reduces attention-related complexity to $O(N_c^\alpha d)=O(N^\alpha d)$ for 
$\alpha \in (1, 2]$. \rebuttalrevision{A more thorough discussion of subsequence selection can be found in~\cref{sec:app-subsequence-selection}.}  
We remark that applying SL to training leads to a cost-effective system design, as training generally involves a significantly higher computational cost compared to inference.

\cref{tbl:sparsity_30d} presents the \textit{sparsity} (see~\cref{sec:app-stochastic-length}) for different sequence lengths and $\alpha$ values,
for a representative industry-scale configuration with 30-day user 
history. The settings that result in negligible regression in model quality
are underlined and highlighted in blue.
The rows labeled 
``$\alpha=2.0$''
represents the base sparsity case where SL is not applied. Lower $\alpha$'s are applicable to longer sequences up to the longest sequence
length we tested, 8,192. 
\begin{table}[t]
\small
\vspace{-.5em}
\begin{center}
\begin{tabular}{crrrr}
\toprule
\multirow{2}{*}{\bf{Alpha} ($\alpha$)} & \multicolumn{4}{c}{\bf Max Sequence Lengths} \\
& 1,024 & 2,048 & 4,096 & 8,192  \\
\midrule
1.6 & 71.5\% & 76.1\% & 80.5\% & \underline{\color{blue}{84.4\%}} \\
1.7 & \underline{\color{blue}{56.1\%}} & \underline{\color{blue}{63.6\%}} & \underline{\color{blue}{69.8\%}} & \underline{\color{blue}{75.6\%}} \\
1.8 & \underline{\color{blue}{40.2\%}} & \underline{\color{blue}{45.3\%}} & \underline{\color{blue}{54.1\%}} & \underline{\color{blue}{66.4\%}} \\
1.9 & \underline{\color{blue}{17.2\%}} & \underline{\color{blue}{21.0\%}} & \underline{\color{blue}{36.3\%}} & \underline{\color{blue}{64.1\%}} \\
2.0 & \underline{\color{blue}{3.1\%}} & \underline{\color{blue}{6.6\%}} & \underline{\color{blue}{29.1\%}} & \underline{\color{blue}{64.1\%}} \\
\bottomrule
\end{tabular}
\end{center}
\vspace{-.8em}
\caption{Impact of \textit{Stochastic Length} (SL) on sequence sparsity.} 
\label{tbl:sparsity_30d}
\vspace{-1.9em}
\end{table}

\vspace{-.5em}
\subsection{Minimizing activation memory usage}
\label{sec:design_activation_memory}
In recommendation systems, the use of large batch sizes is crucial for both
training throughput~\citep{dlrm_isca22} and
model quality~\citep{mixed_negative_sampling_goog_www20,simclr_icml20,meta23ndp}.
Consequently, activation memory usage becomes a major scaling bottleneck,
in contrast to large language models that are commonly trained with small batch sizes and dominated by parameter memory usage.

Compared to Transformers, HSTU employs a simplified and fully fused design that significantly reduces activation memory usage. Firstly, HSTU reduces the number of linear layers outside of attention from six to two,
aligning with recent work that uses elementwise gating to reduce MLP computations~\citep{icml2022flash,s4_iclr22}.
Secondly, HSTU aggressively fuses computations into single operators, including 
$\phi_1(f_1(\cdot))$ in~\cref{eq:hstu-pointwise-projection}, and layer norm, optional dropout, and output MLP
in \cref{eq:hstu-pointwise-transformation}.
This simplified design
reduces the activation memory usage to
$2 d + 2 d + 4hd_{qk} + 4hd_v + 2hd_v = 14d$ 
per layer in bfloat16.

For comparison, Transformers use a feedforward layer and dropout after attention (intermediate state of $3 hd_v$), followed by a pointwise feedforward block consisting of layer norm, linear, activation, linear, and dropout, with intermediate states of $2d + 4d_{ff} + 2d + 1d = 4d + 4d_{ff}$. Here, we make standard assumptions that $hd_v \geq d$ and that $d_{ff} = 4 d$~\citep{transformers_goog_neurips17,brown2020language}. Thus, after accounting for input and input layer norm ($4d$) and qkv projections, the total activation states is 
$33 d$. HSTU's design hence enables scaling to $>2$x deeper layers. 


Additionally, large scale atomic ids used to represent vocabularies also require significant memory usage. With a 10b vocabulary, 512d embeddings, and Adam optimizer, storing embeddings and optimizer states in fp32 already requires 60TB memory. 
To alleviate memory pressure, 
we employ rowwise AdamW optimizers~\citep{rowwiseadagrad_jmlr14,fbgemm21arxiv} and place optimizer states on DRAM, 
which reduces HBM usage per float from 12 bytes to 2 bytes.

\subsection{Scaling up inference via cost-amortization}
\label{sec:design_e2e_inference}
The last challenge we address is the large number of candidates recommendation systems need to process at serving time. 
\rebuttalrevision{We focus on ranking as for retrieval, encoder cost is fully amortizable, and efficient algorithms exist for both MIPS leveraging quantization, hashing, or partitioning~\citep{pq_nns_pami11,alsh_ping_neurips2014,clusteringss_tkde02,hdss_sigmod11} and non-MIPS cases via beam search or hierarchical retrieval~\citep{otm_icml20,meta23ndp}.} 

For ranking, we have up to tens of thousands of candidates~\citep{ytdnn_goog_recsys16,wang2020cold_preranking}.
We propose an algorithm M-FALCON (Microbatched-Fast Attention Leveraging Cacheable OperatioNs) 
to perform inference for $m$ candidates with an input sequence size of $n$. 

Within a forward pass, M-FALCON handles $b_m$ candidates in parallel by modifying attention masks and $\text{rab}^{p,t}$ biases such that the attention operations performed for $b_m$ candidates are exactly the same. This reduces the cost of applying cross-attention from $O(b_m n^2d)$ to $O((n + b_m)^2d) = O(n^2d)$ when $b_m$ can be considered a small constant relative to $n$. We optionally divide the overall $m$ candidates into $\lceil m / b_m\rceil$ microbatches of size $b_m$ to leverage encoder-level KV caching~\citep{pope2022scalinginference} either across forward passes to reduce cost, or across \textit{requests} to minimize tail latency \rebuttalrevision{(More detailed discussions in~\cref{sec:app-m-falcon})}.

Overall, M-FALCON enables model complexity to linearly scale up with the number of candidates 
in traditional DLRMs's ranking stages; \rebuttalrevision{we succeeded in applying \textit{a 285x} 
more complex target-aware cross attention model at 1.5x-3x throughput} with a constant inference budget for a typical ranking
configuration discussed in~\cref{sec:exp-gr}.

%% file: sec_evaluations_v2.tex
\vspace{-.5em}
\section{Experiments}
\label{sec:experiments}


\subsection{Validating Inductive Hypotheses of HSTU Encoder}
\label{sec:exp-hstu-encoder}

\begin{table*}[t]
\small
\vspace{-1.5em}
\caption{Evaluations of methods on public datasets in multi-pass, full-shuffle settings.}
\label{tbl:public-data}
\begin{center}
  \begin{tabular}{cllllllll}
    \toprule
                        & \bf Method       & \bf HR@10           & \bf HR@50        &  \bf HR@200     & \bf NDCG@10 & \bf NDCG@200\\
       \midrule
\multirow{3}{*}{ML-1M} & SASRec (2023)  & .2853                       & .5474                      & .7528              & .1603           & .2498 \\
                       & HSTU           & .3097 (+8.6\%)              & .5754 (+5.1\%)             & .7716 (+2.5\%)     & .1720 (+7.3\%) & .2606 (+4.3\%) \\
                       & HSTU-large     & \bf .3294 (+15.5\%) & \bf .5935 (+8.4\%) & \bf .7839 (+4.1\%) & \bf .1893 (+18.1\%) & \bf .2771 (+10.9\%) \\
       \midrule
\multirow{3}{*}{ML-20M} & SASRec (2023)  & .2906                      & .5499                       & .7655              & .1621                               & .2521 \\
                        & HSTU           & .3252 (+11.9\%)           & .5885 (+7.0\%)              & .7943 (+3.8\%)     & .1878 (+15.9\%)                     & .2774 (+10.0\%) \\
                        & HSTU-large    & \bf .3567 (+22.8\%) & \bf .6149 (+11.8\%) & \bf .8076 (+5.5\%) & \bf .2106 (+30.0\%) & \bf .2971 (+17.9\%) \\
       \midrule
\multirow{3}{*}{Books} & SASRec (2023) & .0292              & .0729               & .1400               & .0156                 & .0350\\
                       & HSTU          & .0404 (+38.4\%)     & .0943 (+29.5\%)     & .1710 (+22.1\%)     & .0219 (+40.6\%)        & .0450 (+28.6\%) \\
                       & HSTU-large    & \bf .0469 (+60.6\%) & \bf .1066 (+46.2\%) & \bf .1876 (+33.9\%) & \bf .0257 (+65.8\%)    & \bf .0508 (+45.1\%) \\
  \bottomrule
\end{tabular}
\end{center}
\vspace{-1.5em}
\end{table*}



\subsubsection{Traditional Sequential Settings}
\label{sec:exp-traditional-sequential-recommdenders}
We first evaluate the performance of HSTU on two popular recommender datasets, MovieLens and Amazon Reviews. We follow sequential recommendation settings in literature, including \textit{full shuffle} and \textit{multi-epoch} training. For baseline, we use SASRec, a state-of-the-art Transformer implementation~\citep{sasrec_icdm18}~\footnote{\rebuttalrevision{Results for other baselines are reported in~\cref{sec:appendix-eval-traditional-seq-rec}.}}. We report Hit Rate@K and NDCG@K over the entire corpus, consistent with recent work~\citep{seqreceval_recsys21,meta23ndp}. 

Results are presented in~\cref{tbl:public-data}. ``SASRec (2023)'' denotes the best SASRec recipe reported in~\citep{meta23ndp}. The rows labeled 
``HSTU'' use identical configurations as SASRec (same number of layers, heads, etc.). ``HSTU-large'' represents larger HSTU encoders (4x number of layers and 2x number of heads). Results show that a) HSTU, with its design optimized for recommendations, 
significantly outperforms the baseline when using the same configuration, and b) HSTU further improves performance when scaled up. 

It is important to note that the evaluation methodology used here differs significantly from industrial-scale settings, as \emph{full-shuffle} and \emph{multi-epoch} training are generally not practical in streaming settings used in industry~\citep{bd2022monolith}.

\subsubsection{Industrial-scale Streaming Settings}
\label{sec:exp-encoder-quality-industrial-streaming}
We next compare the performance of HSTU, ablated HSTUs, and transformers using
industrial-scale datasets in a 
streaming setting.  Throughout the rest of this section, we report 
Normalized Entropy (NE)~\citep{metaads} for ranking.  We train the models over 100B examples (DLRM equivalent), 
with 64-256 H100s used per job. Given ranking is done in a multi-task setting, we report the main engagement event (``E-Task'') and the main consumption event (``C-Task'').
In our 
context, we consider a 0.001 reduction in NE significant as it generally leads to .5\% topline metric improvements for billions of users. For retrieval, given the setup is similar to language modeling, we report log perplexity. 
We fix encoder parameters in a smaller-scale setting ($l=3$, $n=2048$, $d=512$ for ranking and $l=6$, $n=512$, $d=256$ for retrieval), and grid-search other hyperparameters due to resource limits.

%
\begin{table}[t]
\small
\vspace{-1.2em}
\caption{Evaluation of HSTU, ablated HSTU, and Transformers on industrial-scale datasets in one-pass streaming settings.}
\vspace{-.2em}
\label{tbl:industry-data}
\begin{center}
\begin{tabular}{lrrr}
\toprule
\multicolumn{1}{c}{\multirow{2}{*}{\bf Architecture}}  &\multicolumn{1}{c}{\bf Retrieval} &\multicolumn{2}{c}{\bf Ranking (NE)} \\
    &  log pplx. &     E-Task & C-Task   \\
\hline
        Transformers                        & 4.069              & NaN        & NaN    \\
        HSTU (-$\text{rab}^{p,t}$, Softmax) & 4.024              & .5067      & .7931  \\ 
        HSTU (-$\text{rab}^{p,t}$)          & 4.021              & .4980      & .7860  \\
        Transformer++                       & 4.015              & .4945      & .7822  \\
        HSTU (original rab)                 & 4.029             & .4941      & .7817  \\
        HSTU                                & \bf 3.978 & \bf .4937 & \bf .7805 \\
\bottomrule
\end{tabular}
\end{center}
\vspace{-1.5em}
\end{table}

We show results in~\cref{tbl:industry-data}. 
First, HSTU significantly outperforms Transformers, especially in ranking, likely due to pointwise attention and improved relative attention biases.
Second, the gaps between the ablated HSTUs and HSTU confirm the effectiveness of our designs. Optimal learning rates are about 10x lower for Softmax-based HSTU and Transformer vs the rest due to training stability. Even with lower learning rates and pre-norm residual 
connections~\citep{transformer_ln_icml20}, we encountered frequent loss explosions with standard Transformers 
in ranking. 
Finally, HSTU outperforms a popular Transformer variant used in LLMs, Transformer++~\citep{touvron2023llama1}, which uses RoPE~\citep{su2023roformer}, SwiGLU, etc. 
Overall, in this small-scale setting,
HSTU shows better quality at 1.5x-2x faster wall-clock time and 50\% less HBM usage. 

\subsection{Encoder Efficiency}
\label{sec:exp-encoder-efficiency}

\begin{figure}



    \begin{center}
    \includegraphics[width=0.49\columnwidth]{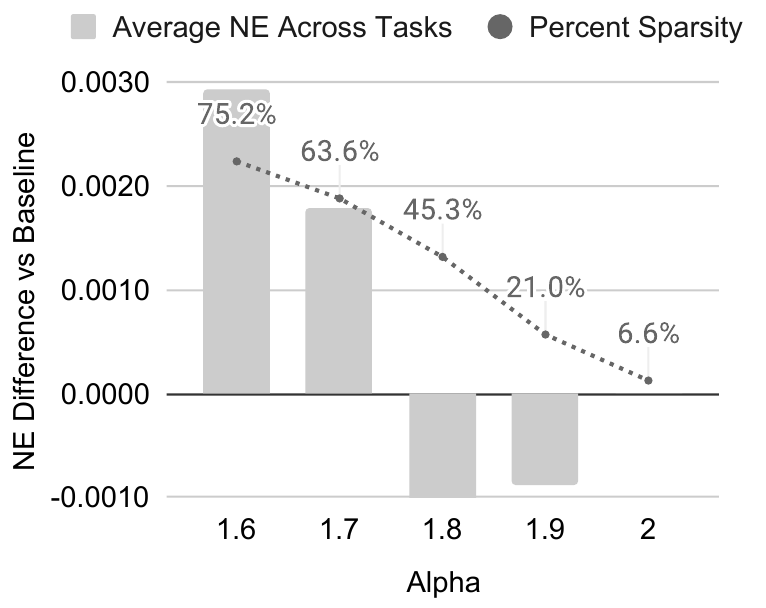}
    \includegraphics[width=0.49\columnwidth]{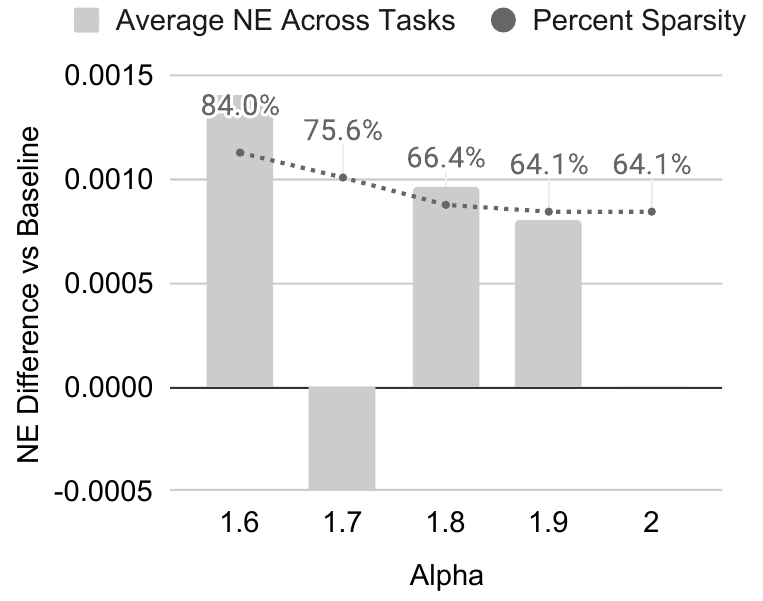}
    \end{center}
    \vspace{-1.6em}
    \caption{Impact of \textit{Stochastic Length} (SL) on metrics. Left: $n=4096$. Right: $n=8192$. Full results can be found in~\cref{sec:app-stochastic-length}.}
    \label{fig:exp_stochastic_length}
    \vspace{-1.5em}
\end{figure}

\begin{figure}[h]
    \vspace{-.8em}
    \begin{center}
    \includegraphics[width=0.75\linewidth]{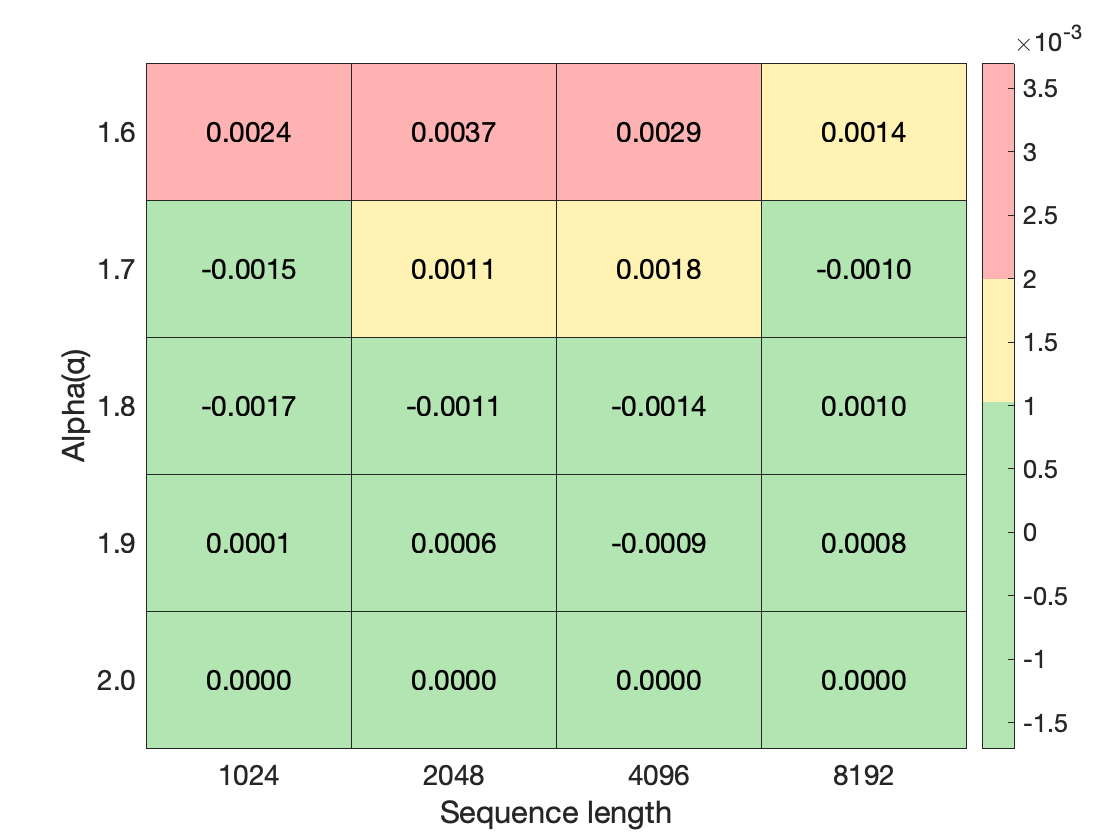}
    \vspace{-.2em}
    \subcaption{Training NE.}
    \vspace{-.2em}
    \includegraphics[width=0.75\linewidth]{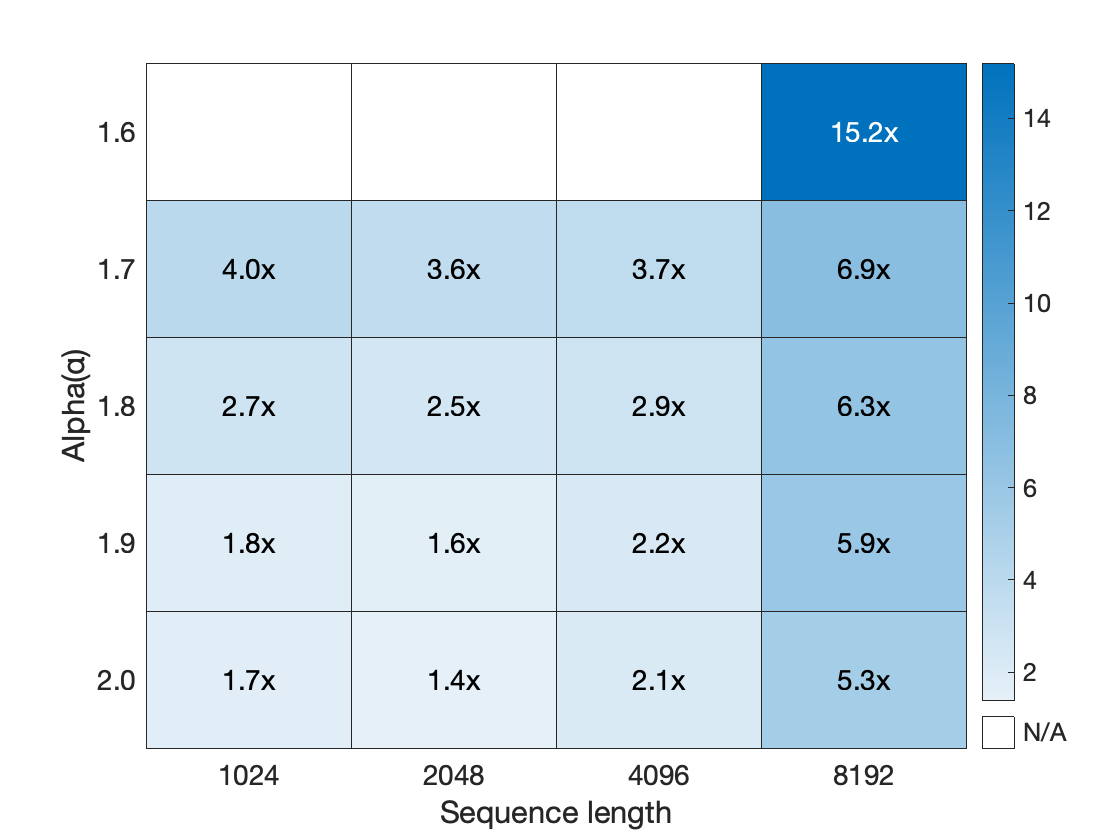}
    \vspace{-.1em}
    \subcaption{Training Speedup.}
    \includegraphics[width=0.75\linewidth]{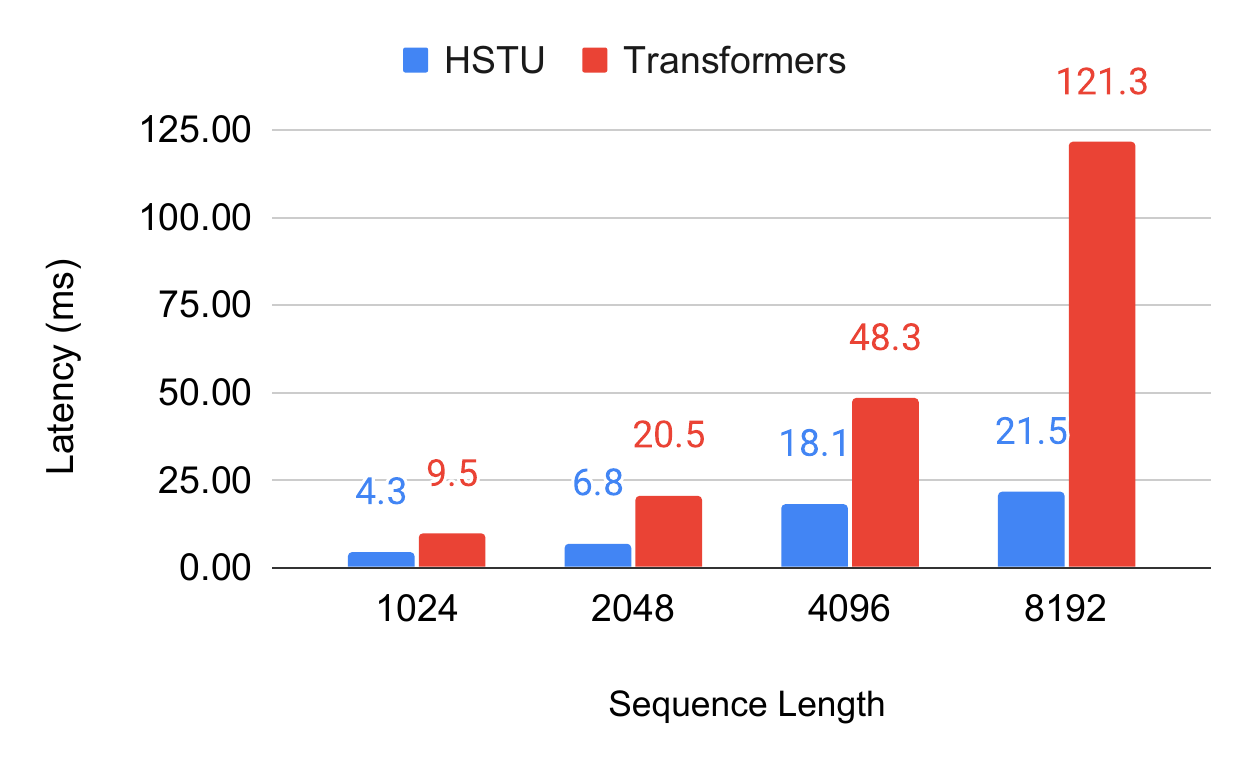}
    \vspace{-.8em}
    \subcaption{Inference Speedup.}
    \end{center}
    \vspace{-1em}
    \caption{\rebuttalrevision{Encoder-level efficiency: HSTU vs FlashAttention2-based Transformers for Training (a, b) and Inference (c).}}
    \label{fig:exp_encoder_efficiency_inference}
    \vspace{-1.9em}
\end{figure}

\textbf{Stochastic Length.} \cref{fig:exp_stochastic_length} and \cref{fig:exp_encoder_efficiency_inference} (a) show the impact of \textit{stochastic length} (SL) on model metrics. At $\alpha=1.6$, a sequence of length $4096$ is turned into a sequence of length $776$ the majority of the time, or removing more than 80\% of the tokens. Even after sparsity ratio increases to 64\%-84\%, the NEs we obtained for main tasks did not degrade by more than 0.002 (0.2\%).
This evidence supports that SL, for suitable $\alpha$s, does not negatively impact model quality and allows for 
high sparsity to reduce training cost.
We further verify in~\cref{sec:app-sequence-length-extrapolation} that SL significantly outperforms existing length extrapolation techniques.

\textbf{Encoder Efficiency.} \cref{fig:exp_encoder_efficiency_inference}
compares the efficiency of
HSTU and Transformer encoders in training and inference settings. 
For Transformers, we use the state-of-the-art
FlashAttention-2~\citep{dao2023flashattention2} implementation.
We consider sequence lengths ranging from 1,024 to 8,192 and apply
\textit{Stochastic Length} (SL) during training.
In the evaluation, we use the same configuration for HSTU and Transformer ($d=512$, $h=8$, $d_{qk}=64$) and ablate \textit{relative attention bias} 
considering HSTU outperforms Transformers without $\text{rab}^{p,t}$, as demonstrated in~\cref{sec:exp-encoder-quality-industrial-streaming}. We compare the encoder-level performance in bfloat16 on NVIDIA H100 GPUs.
Overall, HSTU 
is up to \rebuttalrevision{15.2x} and 5.6x more efficient than Transformers in training and inference, respectively.

Additionally, the decrease in activation memory usage as discussed in Section~\ref{sec:design_activation_memory} enables us to construct over 2x deeper networks with HSTUs compared to Transformers. 

\vspace{-.5em}
\subsection{Generative Recommenders vs DLRMs in Industrial-scale Streaming Settings}
\label{sec:exp-gr}
%
%
Lastly, we compare the end-to-end performance of GRs against state-of-the-art DLRM baselines
in industrial-scale streaming settings. Our GR implementation reflects
a typical configuration used in production, whereas the DLRM settings reflect 
iterations of hundreds of people over multiple years. Given multiple generators are used in the retrieval stage of a recommendation system, we report both the online result for adding GR (``add source'') and replacing existing main DLRM source (``replace source''). 
~\cref{tbl:gr-performance-retrieval} and~\cref{tbl:gr-performance-ranking} show that GR not only significantly outperforms DLRMs offline, but also brings 12.4\% wins in A/B tests.
\begin{table}[t]
\small
\vspace{-1.2em}
\caption{Offline/Online Comparison of Retrieval Models.}
\vspace{-.5em}
\begin{center}
\begin{tabular}{lrrrr}
\toprule
\multicolumn{1}{c}{\multirow{2}{*}{\bf Methods}}  &\multicolumn{2}{c}{\bf Offline HR@K} &\multicolumn{2}{c}{\bf Online metrics}
\\ 
    & K=100 & K=500  &     E-Task & C-Task   
\\ \hline
        DLRM                    &  29.0\%  &  55.5\%  &  +0\%  & +0\%      \\
        DLRM (abl. features)    &  28.3\%  & 54.3\% & -- \\
        GR (content-based)      & 11.6\%  & 18.8\% & -- \\
        GR (interactions only)  & 35.6\%  & 61.7\% & -- \\
        GR (new source)         & \multirow{2}{*}{\bf 36.9\%}  & \multirow{2}{*}{\bf 62.4\%} & \bf +6.2\% & \bf +5.0\% \\
        GR (replace source)        &  &  & \bf +5.1\% & \bf +1.9\%  \\
\bottomrule
\end{tabular}
\end{center}
\label{tbl:gr-performance-retrieval}
\vspace{-1em}
\end{table}

\begin{table}[t]
\small
\vspace{-1.2em}
\caption{Offline/Online Comparison of Ranking Models.}
\vspace{-.5em}
\label{tbl:gr-performance-ranking}
\begin{center}
\begin{tabular}{lrrrr}
\toprule
\multicolumn{1}{c}{\multirow{2}{*}{\bf Methods}}  &\multicolumn{2}{c}{\bf Offline NEs} &\multicolumn{2}{c}{\bf Online metrics}
\\ 
    & E-Task & C-Task  &     E-Task & C-Task   

\\ \hline 
        DLRM                   &  .4982   & .7842       & +0\%                    & +0\%  \\
        \rebuttalrevision{DLRM (DIN+DCN)}      & \rebuttalrevision{.5053}    & \rebuttalrevision{.7899}       & --                      & -- \\
        DLRM (abl. features)   & .5053    & .7925       & --                      & -- \\
        GR (interactions only) & .4851   & .7903  & --  & --  \\
        GR                     & \bf .4845      & \bf .7645      & \bf +12.4\%    & \bf +4.4\%
\\
\bottomrule
\end{tabular}
\end{center}
\vspace{-1.8em}
\end{table}
As discussed in~\cref{sec:dlrms-to-grs}, GRs build upon raw categorical engagement features,
while DLRMs are typically trained with a significantly larger number of features, the majority of which
are handcrafted from raw signals. 
If we give the same set of features used in GRs to DLRMs (``DLRM (abl. features)''), the performance of DLRMs is significantly degraded, which suggests GRs can meaningfully capture those features via their architecture and unified feature space.

We further validate the GR formulation in~\cref{sec:dlrms-to-grs-tasks}
by comparing it with a traditional sequential recommender setup that only considers items user interacted with~\citep{sasrec_icdm18} (``GR (interactions only)''). The results are significantly worse, with its ranking variant underperforming GRs by 2.6\% in NE in the main consumption task.

Considering the popularity of content-based methods (including LMs), we also include a GR baseline with only content features (``GR (content-based)''). The substantial gap in performance of content-based baselines and DLRMs/GRs underscores the significance of high cardinality 
user actions. 






\begin{figure}[h]
    \vspace{-.5em}
    \begin{center}
    \includegraphics[width=0.76\linewidth]{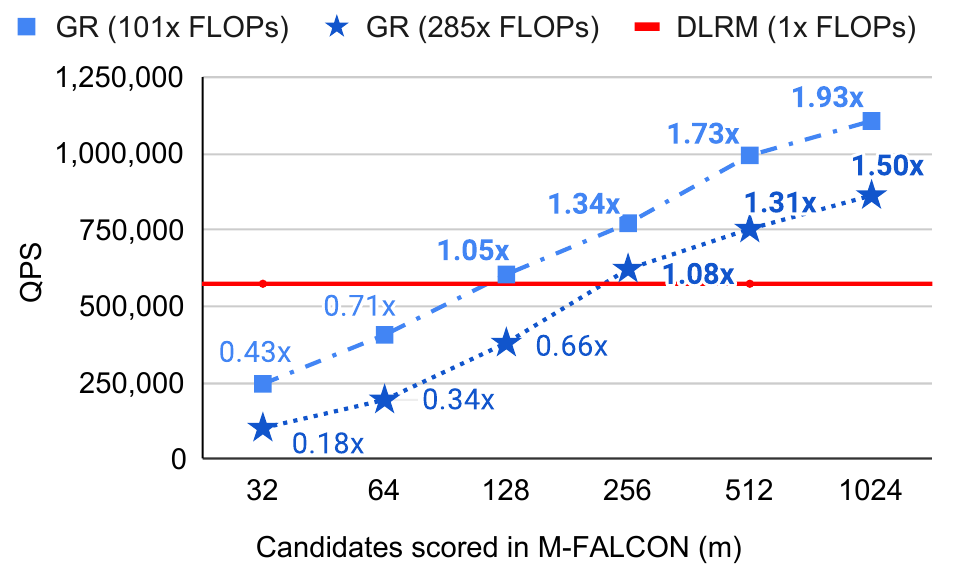}
    \end{center}
    \vspace{-1.5em}
\caption{\rebuttalrevision{Comparison of inference throughput, in the most challenging ranking setup. Full results can be found in~\cref{sec:app-comparison-inference-throughput}.}}
\label{fig:gr-inference-throughput-v2}
\vspace{-1.8em}
\end{figure}

We finally compare the efficiency of GRs with our production DLRMs in~\cref{fig:gr-inference-throughput-v2}.
\rebuttalrevision{Despite the GR model being 285x more computationally complex, we 
achieved 1.50x/2.99x \textit{higher}
QPS when scoring 1024/16384 candidates, }
 due to HSTU and the novel 
M-FALCON 
algorithm in~\cref{sec:design_e2e_inference}. 

\vspace{-.3em}
\subsubsection{Scaling Law for Recommendation Systems}
\label{sec:exp-scalability}
\begin{figure}[t]
    \vspace{-.5em}
    \begin{center}
    \includegraphics[width=0.9\linewidth]{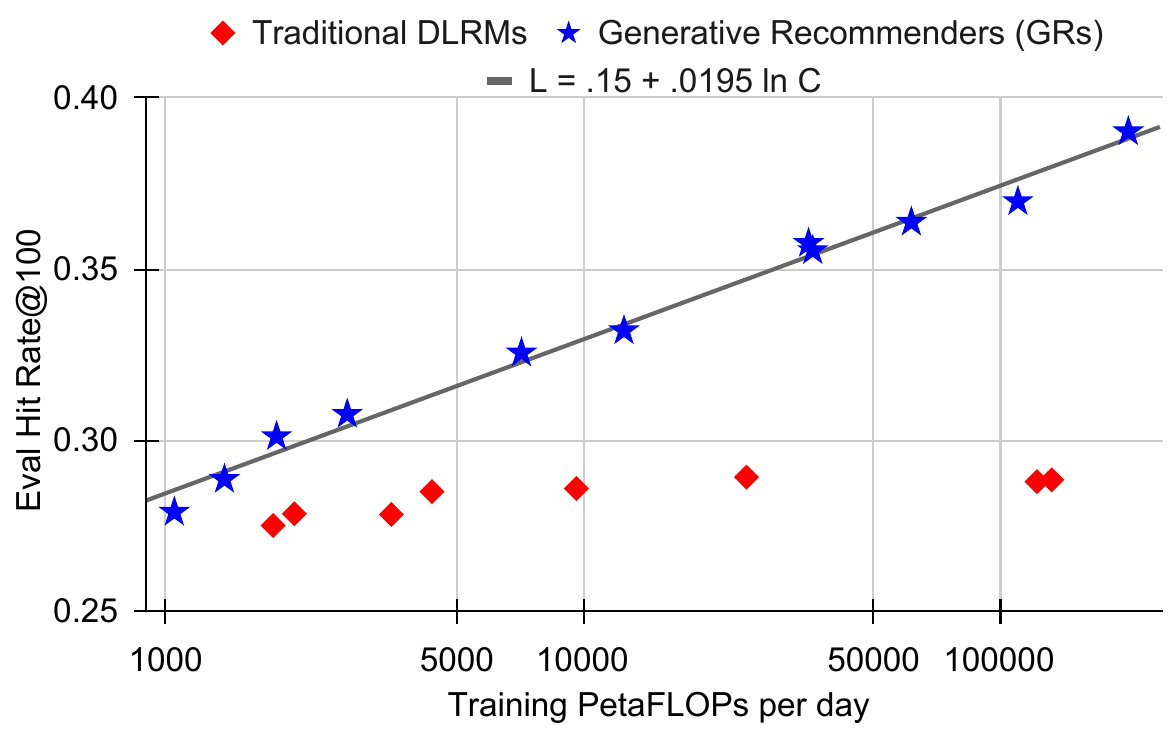}
    \vspace{0.1em}
    \includegraphics[width=0.9\linewidth]{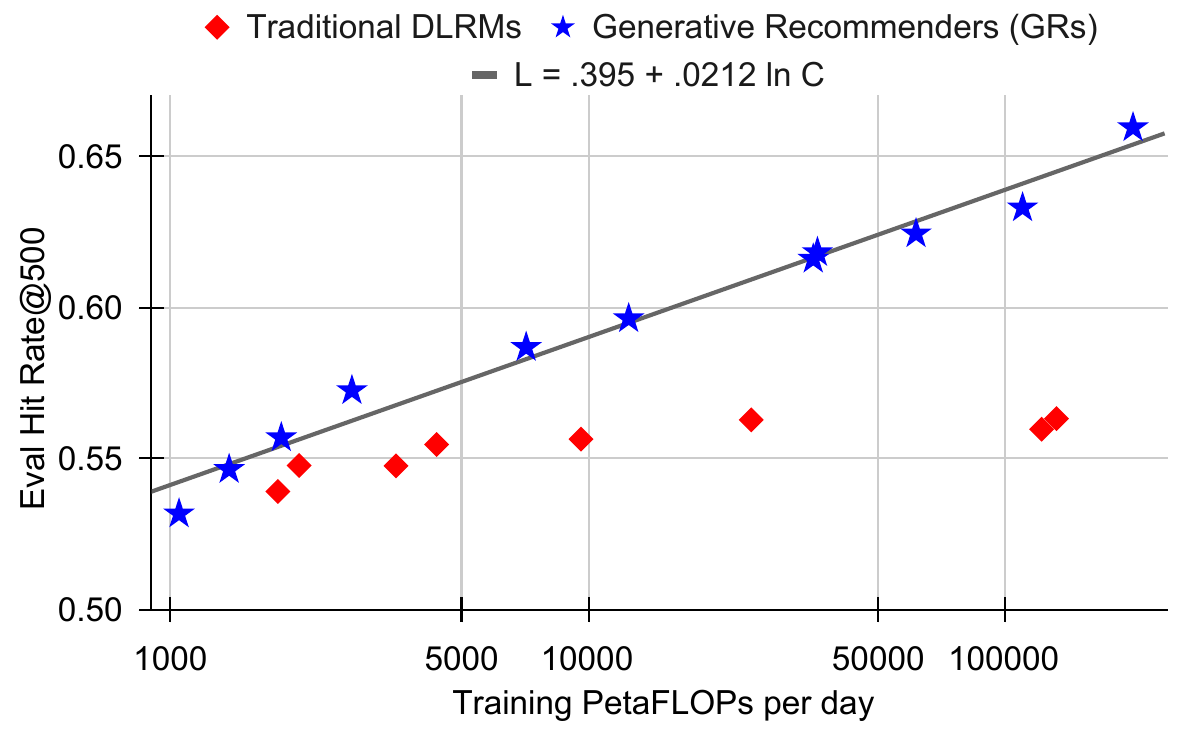}
    \vspace{0.1em}
    \includegraphics[width=0.9\linewidth]{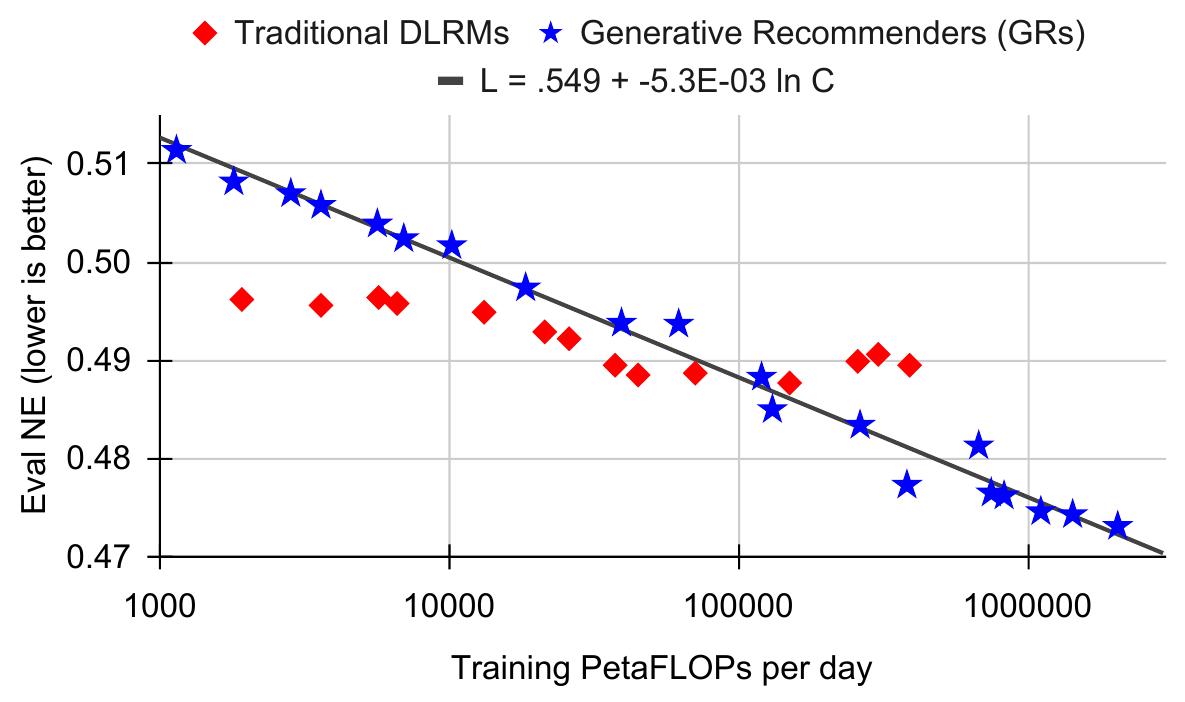}
    \end{center}
    \vspace{-1.7em}
    \caption{\rebuttalrevision{Scalability: DLRMs vs GRs in large-scale industrial settings across retrieval (top, middle) and ranking (bottom). +0.005 in HR and -0.001 in NE represent significant improvements.}} 
    \label{fig:exp_scalability_scaling_law}
    \vspace{-1.75em}
\end{figure}

%

It is commonly known that in large-scale industrial settings, DLRMs saturate in quality at certain compute and params regimes~\citep{zhao2023ucranking}. 
We compare the scalability of GRs and DLRMs to better understand this phenomenon. 

\rebuttalrevision{Since feature interaction layers are crucial for DLRM's performance~\citep{dlrm_isca22}, we experimented with Transformers~\citep{transformers_goog_neurips17}, DHEN~\citep{zhang2022dhen}, and a variant of DCN~\citep{dcnv2_www21} augmented with residual connections~\citep{He2015_resnet} used in our production settings to scale up the DLRM baseline in the ranking setting. For the retrieval baseline, given our baseline used a residual setup, we scaled up hidden layer sizes, embedding dimensions, and number of layers. For HSTU-based Generative Recommenders (GRs), we scaled up the model by adjusting the hyperparameters for HSTU, including the number of residual layers, sequence length, embedding dimensions, number of attention heads, etc. We additionally adjust the number of negatives for retrieval.} 

Results are shown in~\cref{fig:exp_scalability_scaling_law}.
\rebuttalrevision{In the low compute regime, DLRMs might outperform GRs due to handcrafted features, corroborating the importance of feature engineering in traditional DLRMs.
However, GRs demonstrate substantially better scalability with respect to FLOPs, 
whereas DLRM performance plateaus, consistent with findings in prior work. We also observe better scalability w.r.t. both embedding parameters and non-embedding parameters, with GRs leading to 1.5 trillion parameter models, whereas DLRMs performance saturate at about 200 billion parameters.}

Finally, all of our main metrics, including Hit Rate@100 \rebuttalrevision{and Hit Rate@500} for retrieval, and NE for ranking, empirically scale as a power law of compute used given appropriate hyperparameters.  \rebuttalrevision{We observe this phenomenon across three orders of magnitude, up till the largest models we were able to test (8,192 sequence length, 1,024 embedding dimension, 24 layers of HSTU), at which point the total amount of compute we used (normalized over 365 days as we use a standard streaming training setting) is close to the total training compute used by GPT-3~\citep{brown2020language} and LLaMa-2~\citep{touvron2023llama}, as shown in~\cref{fig:intro_dl_models_pf_days}. Within a reasonable range, the exact model hyperparameters play less important roles compared to the total amount of training compute applied. In contrast to language modeling~\citep{kaplan2020scalinglaws}, sequence length play a significantly more important role in GRs, and it's important to scale up sequence length and other parameters 
in tandem.} This is perhaps the most important advantage of our proposed method, as we've shown for the first time that scaling law from LLMs may also apply to large-scale recommendation systems.

%% file: sec_related_work.tex
\vspace{-.5em}
\section{Related Work}

Prior work on sequential recommenders reduces user interactions to a single homogeneous sequence over items~\citep{gru4rec_iclr16,sasrec_icdm18}. 
Industrial-scale applications of sequential approaches are primarily pairwise attention~\citep{din_baba_kdd18} or sequential encoders as part of DLRMs~\citep{bst_dlpkdd19,pinterest23transact}. Multi-stage attention has been explored in lieu of self-attention
to improve efficiency~\citep{kwai2023twin}. Generative approaches that represent ids as a token series have been explored in retrieval~\citep{otm_icml20}. \rebuttalrevision{We give a more extensive discussion of prior work in~\cref{sec:appendix-gr-background}.}




Efficient attention has been a major research focus area 
due to self-attention's $O(n^2)$ scaling factor, with major work like factorized attentions~\citep{sparsetransformers_openai19}, low-rank approximations~\citep{linearattention_icml20}, 
etc. Recently, alternative formulations for sequential transduction settings have been explored~\citep{s4_iclr22,icml2022flash}. HSTU's elementwise gating design, in particular, is inspired by FLASH~\citep{icml2022flash}. Recent hardware-aware formulations have been shown to significantly reduce memory usage~\citep{rabe2021memoryefficientselfattention,korthikanti2022reducing,bytetransformer2022} and give significantly better wallclock time results
~\citep{dao2022flashattention}. 
Length extrapolation enables models trained on shorter sequences to generalize, though most work focuses on finetuning or improving bias mechanisms
~\citep{lengthextrapolation2022iclr}. Our work instead introduces stochasticity in the length dimension, inspired by work on stochasticity in the depth dimension~\citep{huang2016stochasticdepth}.

Interests in large language models (LLMs) have motivated work to treat various recommendation tasks 
as in-context learning~\citep{sileo_zeroshot_lmrec_ecir22}, instruction tuning~\cite{Bao_2023}, or transfer learning~\cite{li23text} on top of pretrained LLMs.
World knowledge embedded in LLMs can be transferred to downstream tasks~\citep{cui2022m6rec} and improve recommendations in zero-shot or few-shot cases. Textual representations of user behavior sequences have also demonstrated good scaling behaviors on medium-scale datasets~\citep{scalingreco2023aaai}.
Most studies of LLMs for recommendation
have been centered around
low-data regimes; in large-scale settings, they have yet to outperform collaborative filtering 
on MovieLens~\citep{hou2024large}. 

%% file: sec_conclusions.tex
\vspace{-.5em}
\section{Conclusions}
\label{sec:conclusions}

We have proposed Generative Recommenders (GRs), a new paradigm that formulates ranking and retrieval as sequential transduction tasks, allowing them to be trained in a generative manner. 
This is made possible by the novel HSTU encoder design, \rebuttalrevision{which is 5.3x-15.2x faster than
state-of-the-art Transformers on 8192 length sequences}, and through the use of 
new training and inference algorithms such as M-FALCON. 
\rebuttalrevision{With GRs, we deployed
models that are \emph{285x} more complex while using \emph{less} inference compute.} GRs and HSTU have led to 12.4\% metric improvements in production
and have shown superior scaling performance compared to traditional
DLRMs.
Our results corroborate that user actions represent an underexplored modality in generative modeling 
-- to echo our title, \textit{``Actions speak louder than words''}.

The dramatic simplification of features in our work
paves the way for the first
foundation models for recommendations, search, and ads by enabling a unified feature space to be used across domains. The fully sequential setup of GRs also enables recommendation to be formulated in an end-to-end, generative setting. Both of these enable recommendation systems to better assist users holistically.


\subsubsection*{Impact Statements} 
We believe that our work has broad positive implications. Reducing reliance of recommendation, search, and ads systems on the large number of heterogeneous features can make these systems much more privacy friendly while improving user experiences. 
Enabling recommendation systems to attribute users' long term outcomes to short-term decisions via fully sequential formulations could reduce the prevalence of content that do not serve users' long term goals (including clickbaits and fake news) across the web, and better align incentives of platforms with user values. Finally, applications of foundation models and scaling law can help reduce carbon footprints incurred with model research and developments needed for recommendations, search, and related use cases.

%% file: sec_acknowledgements.tex
\section*{Acknowledgements}

This work represents the joint efforts of hundreds of people, and would not be possible without work from the following contributors (alphabetical order): Adnan Akhundov, Bugra Akyildiz, Shabab Ayub, Alex Bao, Renqin Cai, Jennifer Cao, Guoqiang Jerry Chen, Lei Chen, Sean Chen, Xianjie Chen, Huihui Cheng, Weiwei Chu, Ted Cui, Shiyan Deng, Nimit Desai, Fei Ding, Francois Fagan, Lu Fang, Liang Guo, Liz Guo, Jeevan Gyawali, Yuchen Hao, Daisy Shi He, Samuel Hsia, Jie Hua, Yanzun Huang, Hongyi Jia, Rui Jian, Jian Jin, Rahul Kindi, Changkyu Kim, Yejin Lee, Fu Li, Hong Li, Shen Li, Wei Li, Zhijing Li, Xueting Liao, Emma Lin, Hao Lin, Jingzhou Liu, Xingyu Liu, Kai Londenberg, Liang Luo, Linjian Ma, Matt Ma, Yun Mao, Bert Maher, Matthew Murphy, Satish Nadathur, Min Ni, Jongsoo Park, Jing Qian, Lijing Qin, Alex Singh, Timothy Shi, Dennis van der Staay, Xiao Sun, Colin Taylor, Shin-Yeh Tsai, Rohan Varma, Omkar Vichare, Alyssa Wang, Pengchao Wang, Shengzhi Wang, Wenting Wang, Xiaolong Wang, Zhiyong Wang, Wei Wei, Bin Wen, Carole-Jean Wu, Eric Xu, Bi Xue, Zheng Yan, Chao Yang, Junjie Yang, Zimeng Yang, Chunxing Yin, Daniel Yin, Yiling You, Keke Zhai, Yanli Zhao, Zhuoran Zhao, Hui Zhang, Jingjing Zhang, Lu Zhang, Lujia Zhang, Na Zhang, Rui Zhang, Xiong Zhang, Ying Zhang, Zhiyun Zhang, Charles Zheng, Erheng Zhong, Xin Zhuang. We would like to thank 
Shikha Kapoor, Rex Cheung, Lana Dam, Ram Ramanathan, Nipun Mathur, Bo Feng, Yanhong Wu, Zhaohui Guo, Hongjie Bai, Wen-Yun Yang, Zellux Wang, Arun Singh, Bruce Deng, Yisong Song, Haotian Wu, Meihong Wang for product support, and Joseph Laria, Akshay Hegde, Abha Jain, Raj Ganapathy for assistance with program management. 
Finally, we would like to thank Ajit Mathews, Shilin Ding, Hong Yan, Lars Backstrom for their leadership support, and insightful discussions with Andrew Tulloch, Liang Xiong, Kaushik Veeraraghavan, and Gaofeng Zhao.

%% file: sec_appendix.tex
\section{Notations}
\label{sec:app-notations}
We summarize key notations used in this paper in~\cref{tbl:table-of-notations} and~\cref{tbl:table-of-notations-continued}.
\begin{table*}[h]
\begin{center}
\begin{tabular}[p]{cl}
\toprule
\bf Symbol & \bf Description \\
\midrule
\multirow{6}{*}{$\Psi_k(t_j)$}  & \multirow{6}{14cm}{The $k$-th training example ($k$ is ordered globally) emitted \rebuttalrevision{by the feature logging system} at time $t_j$. \rebuttalrevision{In a typical DLRM recommendation system, after the user consumes some content $\Phi_i$ (by responding with an action $a_i$ such as skip, video completion and share), the feature logging system joins the tuple $(\Phi_i, a_i)$ with the features used to rank $\Phi_i$, and \textit{emits} $(\Phi_i, a_i, \text{features for }\Phi_i)$ as a training example $\Psi_k(t_j)$. As discussed in~\cref{sec:dlrms-to-grs-generative-training}, DLRMs and GRs deal with different numbers of training examples, with the number of examples in GRs typically being 1-2 orders of magnitude smaller.}} \\
&  \\
&  \\
&  \\
&  \\
&  \\
\midrule
$n_c$ ($n_{c,i})$ & \rebuttalrevision{Number of contents that user has interacted with (of user/sample $i$).} \\
$\Phi_0, \ldots, \Phi_{n_c-1}$ & \rebuttalrevision{List of contents that a user has interacted with, in the context of a recommendation system.} \\
\multirow{3}{*}{$a_0, \ldots, a_{n_c-1}$} & \multirow{3}{14cm}{List of user actions \rebuttalrevision{corresponding to $\Phi_i$s. When all predicted events are binary, each action can be considered a multi-hot vector over (atomic) events such as} like, \rebuttalrevision{share, comment}, image view, \rebuttalrevision{video initialization, video completion, hide, etc.}} \\
     &  \\
     &  \\
\midrule
\multirow{5}{*}{$E,F$} & \multirow{5}{14cm}{Categorical features in DLRMs, in~\cref{fig:design_dlrms_grs_features_training}. \rebuttalrevision{$E_0,E_1$, $\ldots$, $E_7,E_8$, and $F_0,F_1$, $\ldots$, $F_7$ represent transformations of $(\Phi_0, a_0, t_0), \ldots, (\Phi_{n_c-1}, a_{n_c-1}, t_{n_c-1})$ obtained at various points in time via feature extraction (e.g., most recent 10 liked images, most similar 50 urls that the user clicked on in the past compared to the current candidate, etc.). ``merge \& sequentialize'' denotes the (virtual) reverse process of obtaining the raw engagement series $(\Phi_0, a_0, t_0), \ldots, (\Phi_{n_c-1}, a_{n_c-1}, t_{n_c-1})$.}} \\
& \\
& \\
& \\
& \\
\midrule
\multirow{4}{*}{$G,H$} & \multirow{4}{14cm}{Categorical features in DLRMs, in~\cref{fig:design_dlrms_grs_features_training} \rebuttalrevision{that are not related to user-content engagements. These features (e.g., demographics or followed creators) are merged into the main time series (list of contents user engaged with, e.g., $\Phi_0, a_0, \ldots, \Phi_{n_c-1}, a_{n_c-1}$), as discussed in~\cref{sec:dlrms-to-grs-features} and illustrated in~\cref{fig:design_dlrms_grs_features_training}.}} \\
& \\
& \\
& \\
\midrule
\multirow{2}{*}{$n$ ($n_i$)} & \multirow{2}{14cm}{Number of tokens in the sequential transduction task (of user/sample $i$). \rebuttalrevision{While $O(n) = O(n_c)$, $n$ can differ from $n_c$ even without any non-interaction related categorical features; see e.g.,~\cref{tbl:dlrms-to-grs-task-definitions}.}} \\
&  \\
$x_0, \ldots, x_{n-1}$ & List of input tokens in the sequential transduction task. \\
$y_0, \ldots, y_{n-1}$ & List of output tokens in the sequential transduction task. \\
$t_0, \ldots, t_{n-1}$ & List of timestamps corresponding to when $x_0, \ldots, x_{n-1}$ were observed. \\
$\mathbb{X}$, $\mathbb{X}_c$ & Vocabulary of all input/output tokens (\rebuttalrevision{$\mathbb{X}$}) and its content subset \rebuttalrevision{$(\mathbb{X}_c$).} \\ 
$N$, $N_c$ & $\max_i n_i$, \rebuttalrevision{$\max_i n_{c,i}$}. \\
$u_t$     & User representation at time $t$. \\
$s_u(n_i)$, $\hat{s}_u(n_i)$ & Sampling rate for user $i$, \rebuttalrevision{used in generative training (\cref{sec:dlrms-to-grs-generative-training})}. \\
\midrule
$d$       & Model dimension (embedding dimension). \\
$d_{qk}$  & Attention dimension size in HSTU and Transformers. \rebuttalrevision{This applies to $Q(X)$ and $K(X)$ in~\cref{eq:hstu-pointwise-projection}.} \\
$d_v$     & \rebuttalrevision{Value dimension size in HSTU. For Transformers, we typically have $d_{qk}=d_v$.} \\
\multirow{2}{*}{$d_{ff}$}  & \multirow{2}{14cm}{Hidden dimension size in pointwise feedforward layers of Transformers. \rebuttalrevision{HSTU does not utilize feedforward layers; see $U(X)$ below.}} \\ 
    & \\
$h$ & Number of attention heads. \\
\multirow{2}{*}{$l$} & \multirow{2}{14cm}{Number of layers in HSTU. For Transformers, attention and pointwise feedforward layers together constitute a layer.} \\
    &  \\
\bottomrule
\end{tabular}
\vspace{-.5em}
\caption{Table of Notations (continued on the next page).}
\label{tbl:table-of-notations}
\end{center}
\vspace{-1em}
\end{table*}

\begin{table*}[h]
\begin{center}
\begin{tabular}[p]{cl}
\toprule
\bf Symbol & \bf Description \\
\midrule
\multirow{2}{*}{$X$} & \multirow{2}{13cm}{\rebuttalrevision{Input to an HSTU layer. In standard terminology (before batching), $X \in \mathbb{R}^{N \times d}$ assuming we have a input sequence containing $N$ tokens.}} \\
& \\
\multirow{3}{*}{$Q(X), K(X), V(X)$} & \multirow{3}{13cm}{\rebuttalrevision{Query, key, value in HSTU obtained for a given input $X$ based on~\cref{eq:hstu-pointwise-projection}. The definition is similar to $Q$, $K$, and $V$ in standard Transformers. $Q(X), K(X) \in \mathbb{R}^{h \times N \times d_{qk}}$, and $V(X) \in \mathbb{R}^{h \times N\times d_v}$.}} \\
            &   \\
            &   \\
\multirow{2}{*}{$U(X)$}      & \multirow{2}{13cm}{\rebuttalrevision{HSTU uses $U(X)$ to ``gate'' attention-pooled values ($V(X)$) in~\cref{eq:hstu-pointwise-transformation}, which together with $f_2(\cdot)$, enables HSTU to avoid feedforward layers altogether. $U(X) \in \mathbb{R}^{h \times N \times d_v}$.}}  \\
& \\
$A(X)$  & \rebuttalrevision{Attention tensor obtained for input $X$. $A(X) \in \mathbb{R}^{h \times N \times N}$.} \\
$Y(X)$  & \rebuttalrevision{Output of a HSTU layer obtained for the input $X$. $Y(X) \in \mathbb{R}^d$.} \\
\multirow{4}{*}{$\text{Split}(\cdot)$} & \multirow{4}{13cm}{\rebuttalrevision{The operation that splits a tensor into chunks. $\phi_1(f_1(X))) \in \mathbb{R}^{N \times (2 h d_{qk} + 2h d_v)}$ in~\cref{eq:hstu-pointwise-projection}; we obtain $U(X)$, $V(X)$ (both of shape $h \times N \times d_{v}$), $Q(X)$, $K(X)$ (both of shape $h \times N \times d_{qk}$) by splitting the larger tensor (and permuting dimensions) with $U(X), V(X), Q(X), K(X) = \text{Split}(\phi_1(f_1(X)))$.}} \\
& \\
& \\
& \\
\multirow{4}{*}{$\text{rab}^{p,t}$} & \multirow{4}{13cm}{\rebuttalrevision{relative attention bias that incorporates both positional~\citep{t5_jmlr20} and temporal information (based on the time when the tokens are observed, $t_0, \ldots, t_{n-1}$; one possible implementation is to apply some bucketization function to $(t_j - t_i)$ for $(i, j)$). In practice, we share $\text{rab}^{p,t}$ across different attention heads within a layer, hence $\text{rab}^{p,t} \in \mathbb{R}^{1 \times N \times N}$.}} \\
& \\
& \\
& \\
\midrule
$\alpha$    & Parameter controlling sparsity in the \textit{Stochastic Length} algorithm used in HSTU \rebuttalrevision{(\cref{sec:design-algorithmic-sparsity})}. \\ 
$R$ & Register size \rebuttalrevision{on GPUs, in the context of the HSTU algorithm discussed in~\cref{sec:design-algorithmic-sparsity}}. \\
\midrule
$m$         & Number of candidates considered in a recommendation system's ranking stage. \\
$b_m$       & Microbatch size, in the M-FALCON algorithm \rebuttalrevision{discussed in~\cref{sec:design_e2e_inference}}. \\
\bottomrule
\end{tabular}
\vspace{-.5em}
\caption{Table of Notations (continued)}
\label{tbl:table-of-notations-continued}
\end{center}
\vspace{-1em}
\end{table*}

\section{Generative Recommenders: Background and Formulations}
\label{sec:app-generative-recommenders-background-and-formulations}

\rebuttalrevision{Many readers are likely more familiar with classical Deep Learning Recommendation Models (DLRMs)~\citep{dlrm_isca22} given its popularity from YouTube DNN days~\citep{ytdnn_goog_recsys16} and its widespread usage in every single large online content and e-commerce platform~\citep{wdl_goog_dlrs16,din_baba_kdd18,dcnv2_www21,kwai2023twin,pinterest23transact,meta23ndp}. DLRMs operate on top of heterogeneous feature spaces using various neural networks including feature interaction modules~\citep{deepfm_ijcai17,afm_ijcai17,dcnv2_www21}, sequential pooling or target-aware pairwise attention modules~\citep{gru4rec_iclr16,din_baba_kdd18,kwai2023twin} and advanced multi-expert multi-task modules~\citep{mmoe_kdd18,ple_recsys20}. We hence provided an overview of Generative Recommenders (GRs) by contrasting them with classical DLRMs explicitly in~\cref{sec:dlrms-to-grs} and~\cref{sec:design_hstu_encoder}. In this section, we give the readers an alternative perspective starting from the classical sequential recommender literature.}

\subsection{Background: Sequential Recommendations in Academia and Industry}
\label{sec:appendix-gr-background}
\subsubsection{Academic Research (Traditional Sequential Recommender Settings)}
\label{sec:appendix-academic-research}

\rebuttalrevision{
\textbf{Recurrent neural networks (RNNs)} were first applied to recommendation scenarios in GRU4Rec~\citep{gru4rec_iclr16}. \citet{gru4rec_iclr16} considered Gated Recurrent Units (GRUs) and applied them over two datasets, RecSys Challenge 2015~\footnote{\href{http://2015.recsyschallenge.com/}{http://2015.recsyschallenge.com/}} and VIDEO (a proprietary dataset). In both cases, \textit{only positive events (clicked e-commerce items or videos where users spent at least a certain amount of time watching)} were kept as part of the input sequence. We further observe that in a classical industrial-scale two-stage recommendation system setup consisting of retrieval and ranking stages~\citep{ytdnn_goog_recsys16}, the task that~\citet{gru4rec_iclr16} solved primarily maps to the retrieval task.}

\rebuttalrevision{\textbf{Transformers, sequential transduction architectures, and their variants.} Advances in sequential transduction architectures in later years, in particular Transformers~\citep{transformers_goog_neurips17}, have motivated similar advancements in recommendation systems. SASRec~\citep{sasrec_icdm18} first applied Transformers in an autoregressive setting. They considered the \textit{presence} of a review or rating as \textit{positive} feedback, thereby converting classical datasets like Amazon Reviews~\footnote{\href{https://jmcauley.ucsd.edu/data/amazon/}{https://jmcauley.ucsd.edu/data/amazon/}} and MovieLens~\footnote{\href{https://grouplens.org/datasets/movielens/1m/}{https://grouplens.org/datasets/movielens/1m/},~\href{https://grouplens.org/datasets/movielens/20m/}{https://grouplens.org/datasets/movielens/20m/}} \textit{to sequences of positive items}, similar to GRU4Rec. A binary cross entropy loss was employed, where positive target is defined as the next ``positive'' item (recall this is in essence just presence of a review or rating), and negative target is randomly sampled from the item corpus~$\mathbb{X}=\mathbb{X}_c$.}

\rebuttalrevision{Most subsequent research were built upon similar settings as GRU4Rec~\citep{gru4rec_iclr16} and SASRec~\citep{sasrec_icdm18} discussed above, such as BERT4Rec~\citep{bert4rec_cikm19} applying bidirectional encoder setting from BERT~\citep{bert_naacl19}, S3Rec~\citep{s3rec_cikm20} introducing an explicit pre-training stage, and so on.} 

\subsubsection{Industrial Applications as part of Deep Learning Recommendation Models (DLRMs).}

\rebuttalrevision{Sequential approaches, including sequential encoders and pairwise attention modules, have been widely applied in industrial settings due to their ability to enhance user representations as part of DLRMs.  DLRMs commonly use relatively small sequence lengths, such as 20 in BST~\citep{bst_dlpkdd19}, 1,000 in DIN~\citep{din_baba_kdd18}, and 100 in TransAct~\citep{pinterest23transact}. We observe that these are 1-3 orders of magnitude smaller compared with 8,192 in this work (\cref{sec:exp-gr}).}

\rebuttalrevision{Despite using short sequence lengths, most DLRMs can successfully capture long-term user preferences. This can be attributed to two key aspects. First, precomputed user profiles/embeddings~\citep{pinterest23transact} or external vector stores~\citep{kwai2023twin} are commonly used in modern DLRMs, both of which effectively extend lookback windows. Second, a significant number of contextual-, user-, and item-side features were generally employed~\citep{din_baba_kdd18,bst_dlpkdd19,kwai2023twin,pinterest23transact} and various heterogeneous networks, such as FMs~\citep{afm_ijcai17,deepfm_ijcai17}, DCNs~\citep{dcnv2_www21}, MoEs, etc. are used to transform representations and combine outputs.}

\rebuttalrevision{In contrast to sequential settings discussed in~\cref{sec:appendix-academic-research}, all major industrial work  defines loss over (user/request, candidate item) pairs. In the ranking setting, a multi-task binary cross-entropy loss is commonly used. In the retrieval setting, two tower setting~\citep{ytdnn_goog_recsys16} remains the dominant approach. Recent work has investigated representing the next item to recommend as a probability distribution over a sequence of (sub-)tokens, such as 
OTM~\citep{otm_icml20}, and DR~\citep{dr_cikm21} (note that in other recent work, the same setting is sometimes denoted as ``generative retrieval''). They commonly utilize beam search to decode the item from sub-tokens. Advanced learned similarity functions, such as mixture-of-logits~\citep{meta23ndp}, have also been proposed and deployed as an alternative to two-tower setting and beam search given proliferation of modern accelerators such as GPUs, custom ASICs, and TPUs.}

\rebuttalrevision{From a problem formulation perspective, we consider all work discussed above part of DLRMs~\citep{dlrm_isca22} given the model architectures, features used, and losses used differ significantly from academic sequential recommender research discussed in~\cref{sec:appendix-academic-research}. It's also worth remarking that there have been no successful applications of fully sequential ranking settings in industry, especially not at billion daily active users (DAU) scale, prior to this work.}


\subsection{Formulations: Ranking and Retrieval as Sequential Transduction Tasks in Generative Recommenders (GRs)}
\label{sec:app-grs-formulations}

\rebuttalrevision{We next discuss three limitations in the traditional sequential recommender settings and DLRM settings, 
and how Generative Recommenders (GRs) address them from a problem formulation perspective.}

\textbf{Ignorance of features other than user-interacted items.} \rebuttalrevision{Past sequential formulations only consider contents (items) users explicitly interacted with~\citep{gru4rec_iclr16,sasrec_icdm18,bert4rec_cikm19,s3rec_cikm20}, 
while industry-scale recommendation systems prior to GRs are trained over a vast number of features to enhance the representation of users and contents~\citep{ytdnn_goog_recsys16,wdl_goog_dlrs16,
din_baba_kdd18,bst_dlpkdd19, kwai2023twin,pinterest23transact,meta23ndp}. GR addresses this limitation by a) compressing other categorical features and merging them with the main time series, and b) capturing numerical features through cross-attention interaction utilizing a target-aware formulation as discussed in~\cref{sec:dlrms-to-grs-features} and~\cref{fig:design_dlrms_grs_features_training}. We validate this by showing that the traditional ``interaction-only'' formulation that ignores such features degrades model quality significantly; experiment results can be found in the rows labeled ``GR (interactions only)'' in~\cref{tbl:gr-performance-ranking} and~\cref{tbl:gr-performance-retrieval}, where we show utilizing only interaction history led to a 1.3\% decrease in hit rate@100 for retrieval and a 2.6\% NE decrease in ranking (recall a 0.1\% change in NE is significant, as discussed in~\cref{sec:exp-encoder-quality-industrial-streaming,sec:exp-scalability}).}


\textbf{User representations are computed in a target-independent setting.} \rebuttalrevision{A second issue is most traditional sequential recommenders, including GRU4Rec~\citep{gru4rec_iclr16}, SASRec~\citep{sasrec_icdm18}, BERT4Rec~\citep{bert4rec_cikm19}, S3Rec~\citep{s3rec_cikm20}, 
etc. are formulated in a target-independent fashion where for a target item $\Phi_i$, $\Phi_0, \Phi_1, \ldots, \Phi_{i-1}$ are used as encoder input to compute user representations, which is then used to
provide predictions. In contrast, most major DLRM approaches used in industrial settings formulated the sequential modules used in a target-aware fashion, with the ability to incorporate ``target'' (ranking candidate) information into the user representations. These include DIN~\citep{din_baba_kdd18} (Alibaba), BST~\citep{bst_dlpkdd19} (Alibaba), TWIN~\citep{kwai2023twin} (Kwai), and TransAct~\citep{pinterest23transact} (Pinterest).}

\rebuttalrevision{Generative Recommenders (GRs) combines the best of both worlds by \textit{interleaving} the content and action sequences (\cref{sec:dlrms-to-grs-tasks}) to enable applying target-aware attention in causal, autoregressive settings. We categorize and contrast prior work and this work in~\cref{tbl:appendix-formulations-discrimative-vs-generative}~\footnote{Most large-scale industrial recommenders need to be trained in a streaming/single-pass setting due to vast amount of logged data.}.}

\begin{table*}[h!]
\small
\begin{center}
  \begin{tabular}{cllll}
    \toprule
                        & \multirow{2}{2.2cm}{\bf Input for target item $i$}       & \bf \multirow{2}{2.2cm}{Expected output for target item $i$}  & \multirow{2}{*}{\bf Architecture} & \multirow{2}{*}{\bf Training Procedure} \\
                        &   &   &  & \\
       \midrule
       \multirow{2}{*}{GRs} & \multirow{2}{*}{$\Phi_0,a_0, \Phi_1, a_1, \ldots, \Phi_{i}$}   & 
       \multirow{2}{*}{\bf $a_i$ (target-aware)}  & \multirow{2}{*}{Self-attention (HSTU)} & \multirow{2}{3.4cm}{\bf Causal autoregressive (streaming/single-pass)}\\
       &  &  &  & \\
       \midrule
        GRU4Rec & \multirow{2}{*}{$\Phi_0,\Phi_1,\ldots,\Phi_{i-1}$}     & \multirow{2}{*}{$\Phi_i$} & RNNs (GRUs)  & \multirow{2}{3cm}{\textbf{Causal autoregressive} (multi-pass)} \\
        SASRec  &   &   & Self-attention (Transformers)  \\
      \midrule
        BERT4Rec  & \multirow{2}{2.4cm}{$\Phi_0,\Phi_1,\ldots,\Phi_{i-1}$ (at inference time)}    & \multirow{2}{*}{$\Phi_i$}  &  \multirow{2}{*}{Self-attention (Transformers)} & \multirow{2}{*}{Sequential multi-pass~\footnotemark} \\
        S3Rec  &    &  &   &  \\
      \midrule
       DIN & \multirow{3}{2.4cm}{$\Phi_0,\Phi_1,\ldots,\Phi_i$}        & \multirow{4}{2.4cm}{$a_i$ (\textbf{target aware}, implicitly as part of DLRMs)}  & Pairwise attention & \multirow{4}{3cm}{Pointwise (\textbf{generally streaming/single pass})} \\
       BST &         &   & Self-attention (Transformers) &  \\
       TWIN &         &   & Two-stage pairwise attention &  \\
       TransAct & $(\Phi_0,a_0), \ldots, (\Phi_{i-1}, a_{i-1}),\Phi_i$        &   & Self-attention (Transformers) &  \\
  \bottomrule
\end{tabular}
\vspace{-.5em}
\caption{\rebuttalrevision{Comparison of prior work on sequential recommenders and GRs, in the ranking setting, with DLRMs included for completeness.}}
\label{tbl:appendix-formulations-discrimative-vs-generative}
\end{center}
\vspace{-.5em}
\end{table*}

\footnotetext{BERT4Rec leverages multi-pass training with a mixture of Cloze and pointwise (last item) supervision losses; S3Rec utilizes multi-pass training with pre-training and finetuning as two separate stages.}

\textbf{Discriminative formulations restrict applicability of prior sequential recommender work to pointwise settings.} \rebuttalrevision{Finally, traditional sequential recommenders are discriminative by design. Existing sequential recommender literature, including seminal work such as GRU4Rec and SASRec, 
model $p(\Phi_i | \Phi_0, a_0, \ldots, \Phi_{i-1}, a_{i-1})$, or the conditional distribution of the next item to recommend given users' current states. On the other hand, we observe that there are two probabilistic processes in standard recommendation systems, namely \textit{the process of the recommendation system suggesting a content $\Phi_i$ (e.g., some photo or video) to the user}, and \textit{the process of the user reacting to the suggested content $\Phi_i$ via some action $a_i$} (which can be a combination of like, video completion, skip, etc.).}

\begin{figure}[h]
    \begin{center}
    \includegraphics[width=0.49\columnwidth]{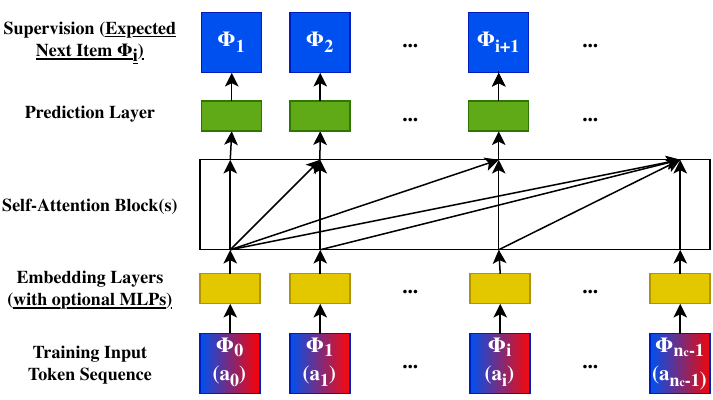}
    \includegraphics[width=0.49\columnwidth]{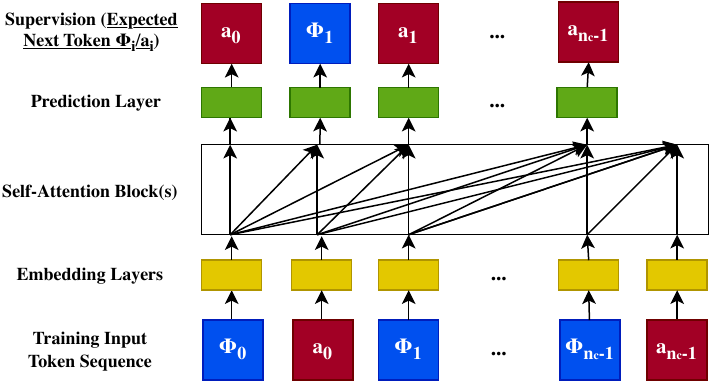}
    \end{center}
    \vspace{-1em}
    \caption{Comparison of traditional sequential recommenders (left) and Generative Recommenders (right). We illustrate sequential recommenders in causal autoregressive settings and GRs without contextual features to facilitate comparison. On the left hand side, the action types $a_i$s are either ignored or combined with item information $\Phi_i$s using MLPs, before going into self-attention blocks.}
    \label{fig:comparison_seqrec_vs_grs}
    \vspace{-0.5em}
\end{figure}

\rebuttalrevision{A generative approach needs to model the joint distribution over the sequence of suggested contents and user actions, or $p(\Phi_0, a_0, \Phi_1, a_1, \ldots, \Phi_{n_c-1}, a_{n_c-1})$, as discussed in~\cref{sec:dlrms-to-grs-tasks}. Our proposal of \textit{Generative Recommenders} enables modeling of such distributions, as shown in~\cref{tbl:dlrms-to-grs-task-definitions-appendix-grs} (\cref{fig:comparison_seqrec_vs_grs}). Note that the next action token ($a_i$) prediction task is exactly the GR ranking setting discussed in~\cref{tbl:dlrms-to-grs-task-definitions}, whereas the next content ($\Phi_i$) prediction task is similar to the retrieval setting adapted to the interleaved setting, with the target changed in order to learn the input data distribution.}

\begin{table}[hb]
\small
\begin{center}
\begin{tabular}{lll}
\toprule
\multicolumn{1}{c}{\bf Task}  & \multicolumn{2}{c}{\bf Specification (Inputs / Outputs / Length)} \\
\hline 
    \multirow{3}{*}{Next action token ($a_i$) prediction} & $x_i$s & $\Phi_0,a_0,\Phi_1,a_1,\ldots,\Phi_{n_c-2},a_{n_c-2},\Phi_{n_c-1},a_{n_c-1}$   \\
                             & $y_i$s  & $a_0,\varnothing,a_1,\varnothing,\ldots,a_{n_c-2},\varnothing,a_{n_c-1},\varnothing$\\
                             & $n$ & $2n_c$ \\
    \midrule
    \multirow{3}{*}{Next content token ($\Phi_i$) prediction} & $x_i$s & $\Phi_0,a_0,\Phi_1,a_1,\ldots,\Phi_{n_c-2},a_{n_c-2},\Phi_{n_c-1},a_{n_c-1}$ \\
                             & \multirow{1}{*}{$y_i$s} & $\varnothing,\Phi_1,\varnothing,\Phi_2,\ldots,\varnothing,\Phi_{n_c-1},\varnothing,\varnothing$  \\
                             & $n$ & $2n_c$ \\
\bottomrule
\end{tabular}
\vspace{-0.5em}
\caption{\rebuttalrevision{Generative modeling over $p(\Phi_0, a_0, \ldots, \Phi_{n_c-1}, a_{n_c-1})$. An illustration is provided in~\cref{fig:comparison_seqrec_vs_grs}.}}
\label{tbl:dlrms-to-grs-task-definitions-appendix-grs}
\end{center}
\vspace{-1em}
\end{table}

\rebuttalrevision{Importantly, this formulation not only enables proper modeling of data distribution but further enables sampling sequences of items to recommend to the user directly via e.g., beam search. We hypothesize that this will lead to a superior approach compared with traditional listwise settings (e.g., DPP~\citep{dpp_neurips14} and RL~\citep{drl_pagewise_recsys18}), and we leave the full formulation and evaluation of such systems (briefly discussed in~\cref{sec:conclusions}) as a future work.}

\section{Evaluation: Synthetic Data}
\label{sec:app-exp-synthetic-data}
As previously discussed in~\cref{sec:pointwise-silu-attention}, standard softmax attention, due to its normalization factor, makes it challenging to capture intensity of user preferences which is important for user representation learning. This aspect is important in recommendation scenarios as the system may need to predict the intensity of engagements (e.g., number of future positive actions on a particular topic) in addition to the relative ordering of items.

To understand this behavior, we construct synthetic data following a Dirichlet Process that generates streaming data over a dynamic set of vocabulary. Dirichlet Process captures the behavior that `rich gets richer` in user engagement histories. We set up the synthetic experiment as follows:

\begin{itemize}
\vspace{-.5em}
    \item We randomly assign each one of 20,000 item ids to exactly one of 100 categories.
\vspace{-.5em}
    \item We generate 1,000,000 records of length 128 each, with the first 90\% being used for training and the final 10\% used for testing. To simulate the streaming training setting, we make the initial 40\% of item ids available initially and the rest available progressively at equal intervals; i.e., at record 500,000, the maximum id that can be sampled is $(40\% + 60\% * 0.5) * 20,000 = 14,000$.
    \item We randomly select up to 5 categories out of 100 for each record and randomly sample a prior $H_c$ over these 5 categories. We sequentially sample category for each position following a Dirichlet process over possible categories as follows:
    \begin{itemize}
        \item for $n > 1$:
        \begin{itemize}
            \item with probability $\alpha / (\alpha + n - 1)$, draw category $c$ from $H_c$.
            \item with probability $n_c / (\alpha + n - 1)$, draw category $c$, where $n_c$ is the number of previous items with category $c$.
            \item randomly sample an assigned item matching category $c$ subject to streaming constraints.
        \end{itemize}
    \end{itemize}
    where $\alpha$ is uniformly sampled at random from $(1.0, 500.0)$.
\end{itemize}

The results can be found in Table~\ref{tbl:synthetic-data}. We always ablate $\text{rab}^{p,t}$ for HSTU as this dataset does not have timestamps. We observe HSTU increasing Hit Rate@10 by more than 100\% relative to standard Transformers. Importantly, replacing HSTU's pointwise attention mechanism with softmax (``HSTU w/ Softmax'') also leads to a significant reduction in hit rate, verifying the importance of pointwise attention-like aggregation mechanisms.

\section{Evaluation: Traditional Sequential Recommender Settings}
\label{sec:appendix-eval-traditional-seq-rec}

\rebuttalrevision{Our evaluations in~\cref{sec:exp-traditional-sequential-recommdenders} focused on comparing HSTU with a state-of-the-art Transformer baseline, SASRec, utilizing latest training recipe. In this section, we further consider two other alternative approaches.}

\rebuttalrevision{\textbf{Recurrent neural networks (RNNs).} We consider the classical work on sequential recommender, GRU4Rec~\citep{gru4rec_iclr16}, to help readers understand how self-attention models, including Transformers and HSTU, compare to traditional RNNs, when all the latest modeling and training improvements are fully incorporated.}

\rebuttalrevision{\textbf{Self-supervised sequential approaches.} We consider the most popular work, BERT4Rec~\citep{bert4rec_cikm19}, to understand how bidirectional self-supervision (leveraged in BERT4Rec via a Cloze objective) compares with unidirectional causal autoregressive settings, such as SASRec and HSTU.}


\begin{table*}[h]
\small
\begin{center}
  \begin{tabular}{cllllllll}
    \toprule
                        & \bf Method       & \bf HR@10           & \bf HR@50        &  \bf HR@200     & \bf NDCG@10 & \bf NDCG@200\\
       \midrule
\multirow{6}{*}{ML-1M} & SASRec (2023)  & .2853                       & .5474                      & .7528              & .1603           & .2498 \\
                       & \rebuttalrevision{BERT4Rec}       & \rebuttalrevision{.2843 (-0.4\%)}             & --                         & --                 & \rebuttalrevision{.1537 (-4.1\%)} & -- \\ 
                       & \rebuttalrevision{GRU4Rec}        & \rebuttalrevision{.2811 (-1.5\%)}              & --                         & --                 & \rebuttalrevision{.1648 (+2.8\%)}  & -- \\
                        & HSTU           & .3097 (+8.6\%)              & .5754 (+5.1\%)             & .7716 (+2.5\%)     & .1720 (+7.3\%) & .2606 (+4.3\%) \\
                       & HSTU-large     & \bf .3294 (+15.5\%) & \bf .5935 (+8.4\%) & \bf .7839 (+4.1\%) & \bf .1893 (+18.1\%) & \bf .2771 (+10.9\%) \\
       \midrule
\multirow{6}{*}{ML-20M} & SASRec (2023)  & .2906                      & .5499                       & .7655              & .1621                               & .2521 \\
                        & \rebuttalrevision{BERT4Rec}       & \rebuttalrevision{.2816 (-3.4\%)}             &  --                          &  -- & \rebuttalrevision{.1703 (+5.1\%)}      &  --      \\
                        & \rebuttalrevision{GRU4Rec}        & \rebuttalrevision{.2813 (-3.2\%)}             &  --                          &  -- & \rebuttalrevision{.1730 (+6.7\%)}      &  --      \\
                        & HSTU           & .3252 (+11.9\%)           & .5885 (+7.0\%)              & .7943 (+3.8\%)     & .1878 (+15.9\%)                     & .2774 (+10.0\%) \\
                        & HSTU-large    & \bf .3567 (+22.8\%) & \bf .6149 (+11.8\%) & \bf .8076 (+5.5\%) & \bf .2106 (+30.0\%) & \bf .2971 (+17.9\%) \\
       \midrule
\multirow{3}{*}{Books} & SASRec (2023) & .0292               & .0729               & .1400               & .0156                 & .0350\\
                       & HSTU          & .0404 (+38.4\%)     & .0943 (+29.5\%)     & .1710 (+22.1\%)     & .0219 (+40.6\%)       & .0450 (+28.6\%) \\
                       & HSTU-large    & \bf .0469 (+60.6\%) & \bf .1066 (+46.2\%) & \bf .1876 (+33.9\%) & \bf .0257 (+65.8\%)   & \bf .0508 (+45.1\%) \\
  \bottomrule
\end{tabular}
\vspace{-0.5em}
\caption{Evaluations of methods on public datasets in traditional sequential recommender settings (multi-pass, full-shuffle). Compared with~\cref{tbl:public-data}, two other baselines (GRU4Rec and BERT4Rec) are included for completeness.}
\label{tbl:public-data-full-appendix}
\end{center}
\vspace{-1em}
\end{table*}

\rebuttalrevision{Results are presented in~\cref{tbl:public-data-full-appendix}.
We reuse BERT4Rec results and GRU4Rec results on ML-1M and ML-20M as reported by~\citet{bert4recvssasrec_recsys23}. 
Given a sampled softmax loss is used, we hold the number of negatives used constant (128 for ML-1M, ML-20M and 512 for Amazon Books) to ensure a fair comparison between methods.}


\rebuttalrevision{The results confirm that SASRec remains one of the most competitive approaches in traditional sequential recommendation settings when sampled softmax loss is used~\citep{meta23ndp,bert4recvssasrec_recsys23}, while HSTU significantly outperforms evaluated transformers, RNNs, and self-supervised bidirectional transformers.}

\section{Evaluation: Traditional DLRM Baselines}
\label{sec:app-dlrm-baselines}


\rebuttalrevision{The DLRM baseline configurations used in~\cref{sec:experiments} reflect continued iterations of hundreds of researchers and engineers over multiple years and a close approximation of production configurations on a large internet platform with billions of daily active users before HSTUs/GRs were deployed. We give a high level description of the models used below.}

\textbf{Ranking Setting.} The baseline ranking model, as described in~\citep{dlrm_isca22}, employs approximately one thousand dense features and fifty sparse features. We incorporated various modeling techniques such as Mixture of Experts~\citep{mmoe_kdd18}, variants of Deep \& Cross Network~\citep{dcnv2_www21}, \rebuttalrevision{various sequential recommendation modules including target-aware pairwise attention (one commonly used variant in industrial settings can be found in~\citep{din_baba_kdd18})}, and residual connection over special interaction layers~\citep{He2015_resnet,zhang2022dhen}. \rebuttalrevision{For the low FLOPs regime in the scaling law section (\cref{sec:exp-scalability}), some modules with high computational costs were simplified and/or replaced with other state-of-the-art variants like DCNs to achieve desired FLOPs.}

\begin{figure}[h]
    \begin{center}
    \includegraphics[width=0.6\linewidth]{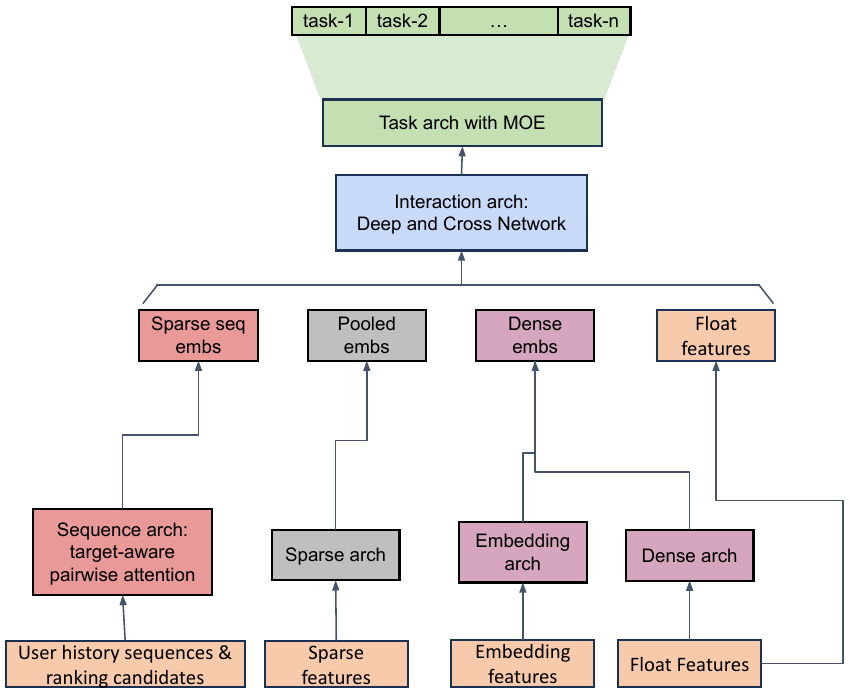}
    \vspace{-.5em}
    \caption{\rebuttalrevision{A high level architecture of a baseline DLRM ranking model (``DLRM (DIN+DCN)'' in~\cref{tbl:gr-performance-ranking}) that utilizes major published work including DIN~\citep{din_baba_kdd18}, DCN~\citep{dcnv2_www21}, and MMoE~\citep{mmoe_kdd18}.}}
    \label{fig:ranking_baseline}
    \end{center}
    \vspace{-0.5em}
\end{figure}

\rebuttalrevision{While we cannot disclose the exact settings due to confidentiality considerations, to the best of our knowledge, our baseline represents one of the best known DLRM approaches when recent research are fully incorporated. To validate this claim and to facilitate readers' understanding, we report a typical setup based on identical features but only utilizing major published results including DIN~\citep{din_baba_kdd18}, DCN~\citep{dcnv2_www21}, and MMoE~\citep{mmoe_kdd18} (``DLRM (DIN+DCN)'') in~\cref{tbl:gr-performance-ranking}, with the combined architecture illustrated in~\cref{fig:ranking_baseline}. This setup significantly underperformed our production DLRM setup by 0.71\% in NE for the main E-Task and 0.57\% in NE for the main C-Task (where 0.1\% NE is significant).}

\textbf{Retrieval Setting.} The baseline retrieval model employs a standard two-tower neutral retrieval setting~\citep{ytdnn_goog_recsys16} with mixed in-batch and out-of-batch sampling. The input feature set consists of both high cardinality sparse features (e.g., item ids, user ids) and low cardinality sparse features (e.g. languages, topics, interest entities). A stack of feed forward layers with residual connections~\citep{He2015_resnet} is used to compress the input features into user and item embeddings.

\rebuttalrevision{\textbf{Features and Sequence Length.} The features used in both of the DLRM baselines, including main user interaction history that is utilized by various sequential encoder/pairwise attention modules, are \textit{strict supersets} of the features used in all GR candidates. This applies to all studies conducted in this paper, including those used in the scaling studies (\cref{sec:exp-scalability}).}


\section{Stochastic Length}
\label{sec:app-stochastic-length}
\subsection{Subsequence Selection}
\label{sec:app-subsequence-selection}
In \cref{eq:sl_sampling}, we select a subsequence of length $L$ from the full user history in order to increase sparsity. Our empirical results indicate that careful design of the subsequence selection technique can improve model quality.
We compute a metric $f_i = t_n - t_i$ which corresponds to the amount of time elapsed since the user interacted with item $x_i$. We conduct offline experiments with the following \rebuttalrevision{subsequence selection methods}: 
\begin{itemize}
    \vspace{-0.2cm}\item \textbf{Greedy \rebuttalrevision{Selection}} -- \rebuttalrevision{Selects} $L$ items with smallest values of $f_i$ from $S$
    \vspace{-0.2cm}\item \textbf{Random \rebuttalrevision{Selection}} -- \rebuttalrevision{Selects} $L$ items from $S$ randomly
    \vspace{-0.2cm}\item \textbf{Feature-Weighted \rebuttalrevision{Selection}} -- \rebuttalrevision{Selects} $L$ items from $S$ according to a weighted distribution $1 - f_{n, i} / (\sum_{j=1}^L f_{j, i}) $
\end{itemize}

During our offline experiments, the feature-weighted \rebuttalrevision{subsequence selection method} resulted in the best model quality,
as shown in~\cref{tbl:sampler_type}. 
\begin{table}[t]
\begin{center}
\begin{tabular}{crrr}
\toprule
\multirow{2}{*}{\bf Metric Name} & \multicolumn{3}{c}{\bf \rebuttalrevision{Selection} Type} \\
& Greedy & Weighted & Random  \\
\midrule
\textbf{Main Engagement Metric (NE)}& 0.495 & 0.494 & 0.495 \\
\textbf{Main Consumption Metric (NE)}& 0.792 & 0.789 & 0.791 \\
\bottomrule
\end{tabular}
\vspace{-.5em}
\caption{Comparison of \rebuttalrevision{subsequence selection methods for \emph{Stochastic Length}} on model quality, measured by Normalized Entropy (NE).}
\label{tbl:sampler_type}
\end{center}
\vspace{-.5em}
\end{table}

\subsection{Impact of Stochastic Length on Sequence Sparsity}
\label{sec:sparsity_appx}
In~\cref{tbl:sparsity_30d}, we show the impact of Stochastic Length on sequence sparsity
for a representative industry-scale configuration with 30-day user engagement history. 
The sequence sparsity is defined as one minus the ratio of the average sequence length of all samples divided by the maximum sequence length. To better characterize the computational cost of sparse attentions,
we also define \textit{s2}, which is defined as one minus the sparsity of the attention matrix.
For reference, we present the results for 60-day and 90-day user engagement history in~\cref{tbl:sparsity_60} and~\cref{tbl:sparsity_90}, respectively.
\begin{table}[t]
\begin{center}
\begin{tabular}{crrrrrrrr}
\toprule
\multirow{2}{*}{\bf Alpha} & \multicolumn{8}{c}{\bf Max Sequence Length} \\
\cline{2-9} \\
& \multicolumn{2}{c}{1,024} & \multicolumn{2}{c}{2,048} & \multicolumn{2}{c}{4,096} & \multicolumn{2}{c}{8,192}  \\
& \multicolumn{1}{c}{sparsity} & \multicolumn{1}{c}{s2} & \multicolumn{1}{c}{sparsity} & \multicolumn{1}{c}{s2} & \multicolumn{1}{c}{sparsity} & \multicolumn{1}{c}{s2} & \multicolumn{1}{c}{sparsity} & \multicolumn{1}{c}{s2} \\
\midrule
1.6 & 71.5\% & 89.4\% & 75.8\% & 92.3\% & 79.4\% & 94.7\% & \underline{\color{blue}{83.8\%}} & \underline{\color{blue}{97.3\%}} \\
1.7 & \underline{\color{blue}{57.3\%}} & \underline{\color{blue}{77.6\%}} & \underline{\color{blue}{60.6\%}} & \underline{\color{blue}{79.8\%}} & \underline{\color{blue}{67.3\%}} & \underline{\color{blue}{86.6\%}} & \underline{\color{blue}{74.5\%}} & \underline{\color{blue}{93.3\%}} \\
1.8 & \underline{\color{blue}{37.5\%}} & \underline{\color{blue}{56.2\%}} & \underline{\color{blue}{42.6\%}} & \underline{\color{blue}{62.1\%}} & \underline{\color{blue}{51.9\%}} & \underline{\color{blue}{74.2\%}} & \underline{\color{blue}{62.6\%}} & \underline{\color{blue}{85.5\%}} \\
1.9 & \underline{\color{blue}{15.0\%}} & \underline{\color{blue}{25.2\%}} & \underline{\color{blue}{17.7\%}} & \underline{\color{blue}{29.0\%}} & \underline{\color{blue}{29.6\%}} & \underline{\color{blue}{47.5\%}} & \underline{\color{blue}{57.8\%}} & \underline{\color{blue}{80.9\%}} \\
2.0 & \underline{\color{blue}{1.2\%}} & \underline{\color{blue}{1.7\%}} & \underline{\color{blue}{2.5\%}} & \underline{\color{blue}{3.5\%}} & \underline{\color{blue}{18.9\%}} & \underline{\color{blue}{30.8\%}} & \underline{\color{blue}{57.6\%}} & \underline{\color{blue}{80.6\%}} \\
\bottomrule
\end{tabular}
\vspace{-.5em}
\caption{Impact of \emph{Stochastic Length} (SL) on sequence sparsity, over a 60d user engagement history.}
\label{tbl:sparsity_60}
\end{center}
\vspace{-1em}
\end{table}
\begin{table}[t]
\begin{center}
\begin{tabular}{crrrrrrrr}
\toprule
\multirow{2}{*}{\bf Alpha} & \multicolumn{8}{c}{\bf Max Sequence Length} \\
\cline{2-9} \\
& \multicolumn{2}{c}{1,024} & \multicolumn{2}{c}{2,048} & \multicolumn{2}{c}{4,096} & \multicolumn{2}{c}{8,192}  \\
& \multicolumn{1}{c}{sparsity} & \multicolumn{1}{c}{s2} & \multicolumn{1}{c}{sparsity} & \multicolumn{1}{c}{s2} & \multicolumn{1}{c}{sparsity} & \multicolumn{1}{c}{s2} & \multicolumn{1}{c}{sparsity} & \multicolumn{1}{c}{s2} \\
\midrule
1.6 & 68.0\% & 85.0\% & 74.6\% & 90.8\% & 78.6\% & 93.5\% & \underline{\color{blue}{83.5\%}} & \underline{\color{blue}{97.3\%}} \\
1.7 & \underline{\color{blue}{56.3\%}} & \underline{\color{blue}{76.1\%}} & \underline{\color{blue}{61.2\%}} & \underline{\color{blue}{80.6\%}} & \underline{\color{blue}{67.5\%}} & \underline{\color{blue}{87.0\%}} & \underline{\color{blue}{74.3\%}} & \underline{\color{blue}{93.3\%}} \\
1.8 & \underline{\color{blue}{38.9\%}} & \underline{\color{blue}{58.3\%}} & \underline{\color{blue}{42.0\%}} & \underline{\color{blue}{61.3\%}} & \underline{\color{blue}{50.4\%}} & \underline{\color{blue}{72.4\%}} & \underline{\color{blue}{61.0\%}} & \underline{\color{blue}{84.4\%}} \\
1.9 & \underline{\color{blue}{16.2\%}} & \underline{\color{blue}{27.3\%}} & \underline{\color{blue}{17.3\%}} & \underline{\color{blue}{28.6\%}} & \underline{\color{blue}{27.2\%}} & \underline{\color{blue}{44.4\%}} & \underline{\color{blue}{54.3\%}} & \underline{\color{blue}{77.8\%}}\\
2.0 & \underline{\color{blue}{0.9\%}} & \underline{\color{blue}{1.2\%}} & \underline{\color{blue}{1.6\%}} & \underline{\color{blue}{2.1\%}} & \underline{\color{blue}{13.5\%}} & \underline{\color{blue}{22.5\%}} & \underline{\color{blue}{54.0\%}} & \underline{\color{blue}{77.4\%}}\\
\bottomrule
\end{tabular}
\vspace{-.5em}
\caption{Impact of \emph{Stochastic Length} (SL) on sequence sparsity, over a 90d user engagement history.}
\label{tbl:sparsity_90}
\end{center}
\end{table}

\begin{figure}[b]
    \vspace{-.8em}    
    \begin{center}
    \includegraphics[width=0.24\linewidth]{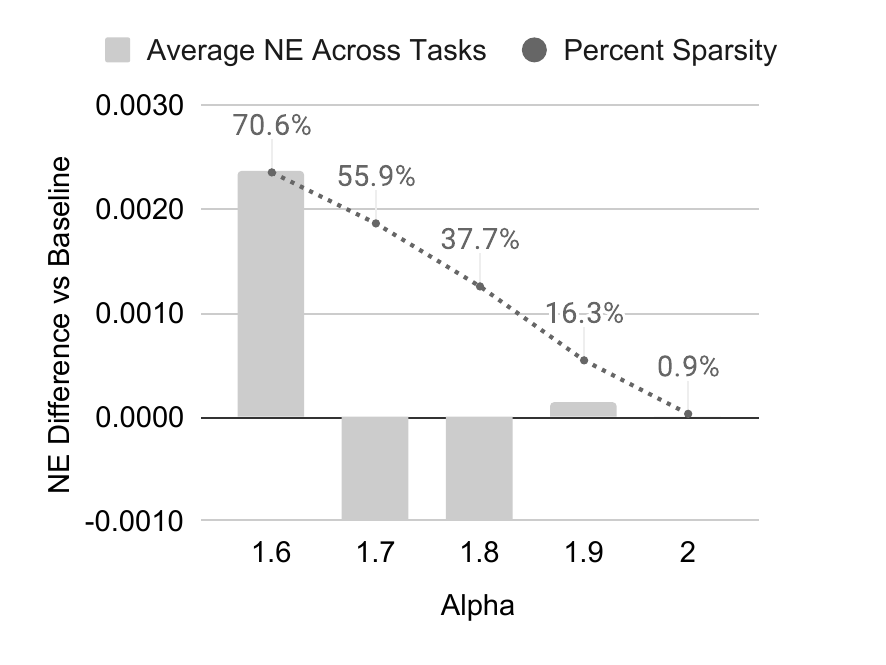}
    \includegraphics[width=0.24\linewidth]{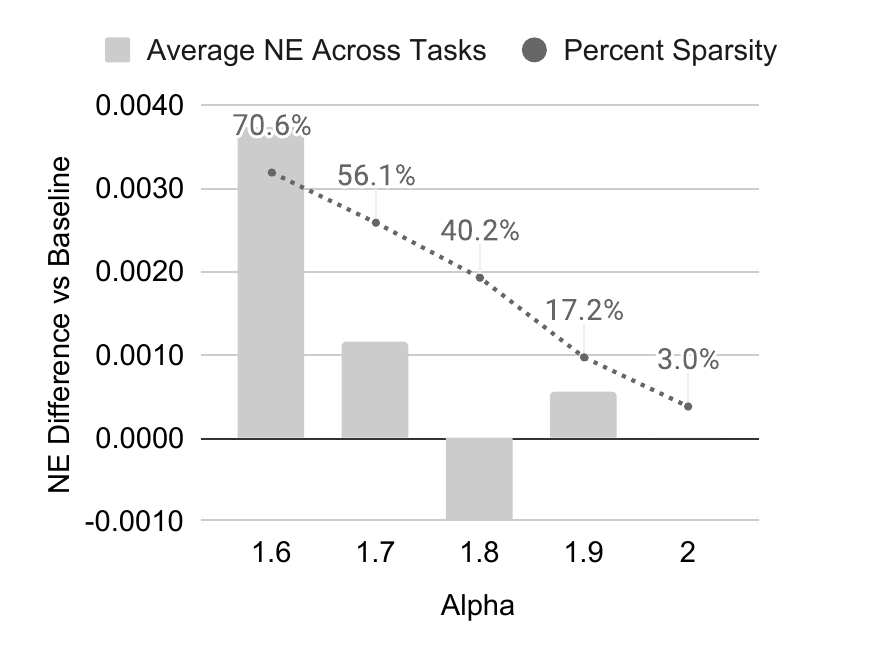}
    \includegraphics[width=0.24\linewidth]{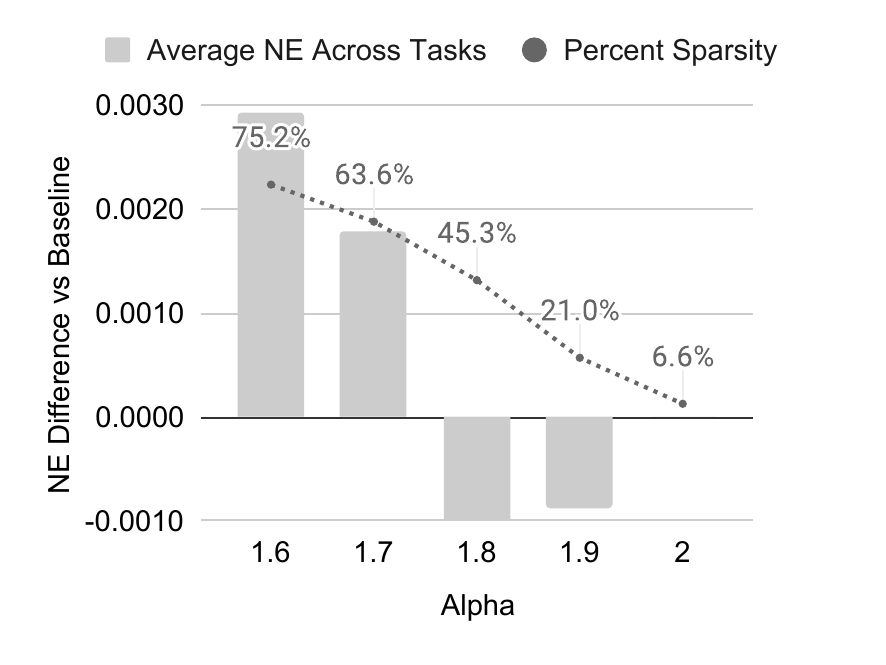}
    \includegraphics[width=0.24\columnwidth]{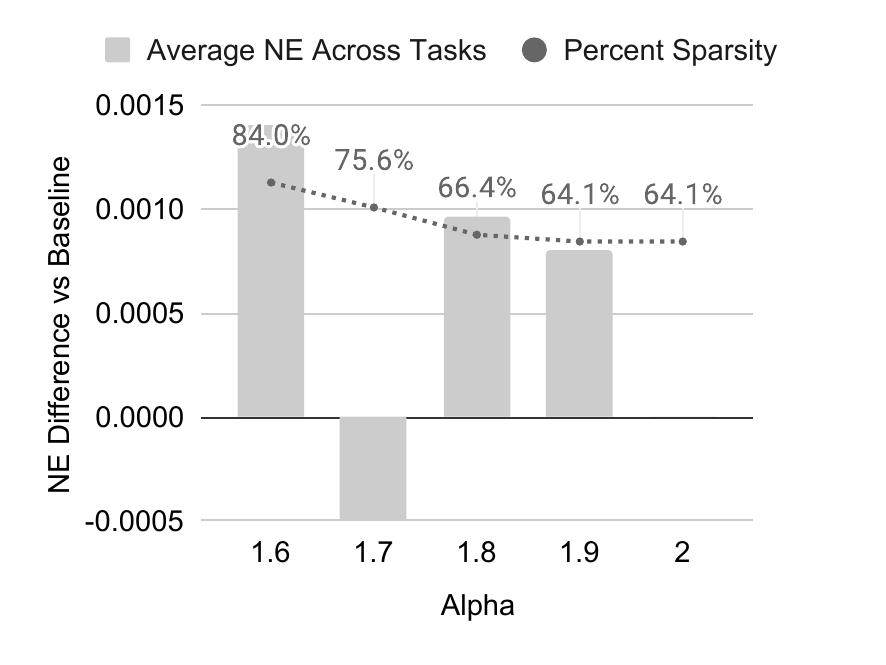}
    \end{center}
    \vspace{-1.5em}
    \caption{Impact of \textit{Stochastic Length} (SL) on ranking model metrics. Left to right: $n = [1024, 2048, 4096, 8192]$ ($n$ is after interleaving algorithm as discussed in~\cref{sec:dlrms-to-grs-tasks} to enable target-aware cross attention in causal-masked settings).}
    \label{fig:app_exp_stochastic_length}
\end{figure}

\subsection{Comparisons Against Sequence Length Extrapolation Techniques}
\label{sec:app-sequence-length-extrapolation}
We conduct additional studies to verify that \textit{Stochastic Length} is competitive against existing techniques for sequence length extrapolation used in language modeling. Many existing methods perform sequence length extrapolation through modifications of RoPE~\citep{su2023roformer}. To compare against existing methods, we train an HSTU variant (HSTU-RoPE) with no relative attention bias and rotary embeddings.

We evaluate the following sequence length extrapolation methods on HSTU-RoPE: 

\begin{itemize}
    \vspace{-0.2cm}\item \textbf{Zero-Shot} - Apply NTK-Aware RoPE~\citep{peng2024yarn} before directly evaluating the model with no finetuning;
    \vspace{-0.2cm}\item \textbf{Fine-tune} - Finetune the model for 1000 steps after applying NTK-by-parts~\citep{peng2024yarn}.
\end{itemize}

We evaluate the following sequence length extrapolation methods on HSTU (includes relative attention bias, no rotary embeddings): 
\begin{itemize}
    \vspace{-0.2cm}\item \textbf{Zero-Shot} - Clamp the relative position bias according to the maximum training sequence length, directly evaluate the model~\citep{t5_jmlr20,lengthextrapolation2022iclr};
    \vspace{-0.2cm}\item \textbf{Fine-tune} - Clamp the relative position bias according to the maximum training sequence length, fine-tune the model for 1000 steps before evaluating the model.
\end{itemize}

In \cref{tbl:seq-len-extrapolation}, we report the NE difference between models with induced data sparsity during training (Stochastic Length, zero-shot, fine-tuning) and models trained on the full data. We define the sparsity for zero-shot and fine-tuning techniques to be the average sequence length during training divided by the max sequence length during evaluation. All zero-shot and fine-tuned models are trained on 1024 sequence length data and are evaluated against 2048 and 4096 sequence length data. In order to find an appropriate \textit{Stochastic Length} baseline for these techniques, we select \textit{Stochastic Length} settings which result in the same data sparsity metrics. 

We believe that zero-shot and fine-tuning approaches to sequence length extrapolation are not well-suited for recommendation scenarios that deal with high cardinality ids. Empirically, we observe that \textit{Stochastic Length} significantly outperforms fine-tuning and zero-shot approaches. We believe that this could be due to our large vocabulary size. Zero-shot and fine-tuning approaches fail to learn good representations for older ids, which could hurt their ability to fully leverage the information contained in longer sequences.

\begin{table*}[t]
\small
\begin{center}
  \begin{tabular}{cccc}
    \toprule
        \multirow{2}{*}{\bf Evaluation Strategy} & \multicolumn{3}{c}{\bf Average NE Difference vs Full Sequence Baseline} \\
        \cline{2-4} \\
        &\multirow{1}{*}{Model Type} & \multicolumn{1}{c}{2048 / 52\% Sparsity} 
        & \multicolumn{1}{c}{ 4096 / 75\% Sparsity} \\
       \midrule
\multirow{2}{*}{Zero-shot} & HSTU \cite{t5_jmlr20} &6.46\% &10.35\% \\
                       & HSTU-RoPE \cite{peng2024yarn}    &7.51\% &11.27\%   \\  
       \midrule
\multirow{2}{*}{Fine-tune} & HSTU \cite{t5_jmlr20} &1.92\%  &2.21\%     \\
                        & HSTU-RoPE \cite{peng2024yarn}    &1.61\% &2.19\%   \\
       \midrule
       \multirow{1}{*}{\textit{Stochastic Length} (SL)} & HSTU  & \textbf{0.098}\%  & \textbf{0.64}\%  \\
  \bottomrule
\end{tabular}
\end{center}
\vspace{-.5em}
\caption{Comparisons of \textit{Stochastic Length} (SL) vs existing Length Extrapolation methods.} 
\label{tbl:seq-len-extrapolation}
\vspace{-0.5em}
\end{table*}

\section{Sparse Grouped GEMMs and Fused Relative Attention Bias}
\label{sec:app-hstu-kernels}
\rebuttalrevision{We provide additional information about the efficient HSTU attention kernel
that was introduced in~\cref{sec:design-algorithmic-sparsity}.
Our approach builds upon Memory-efficient Attention~\citep{rabe2021memoryefficientselfattention} and FlashAttention~\citep{dao2022flashattention}, and is a memory-efficient self-attention mechanism that divides the input into blocks and avoids materializing the large $h \times N \times N$ intermediate attention tensors for the backward pass. By exploiting the sparsity of input sequences,
we can reformulate the attention computation as a group of back-to-back GEMMs
with different shapes. We implement efficient GPU kernels to accelerate this computation. 
The construction of the relative attention bias is also a bottleneck due to memory accesses.
To address this issue, we have fused the relative bias construction and the grouped GEMMs into a single GPU kernel and managed to accumulate gradients using GPU's fast shared memory in the backward pass. Although our algorithm requires recomputing attention and relative bias in the backward pass, it is significantly faster and uses less memory than the standard approach used in Transformers.} 

\newpage
\section{Microbatched-Fast Attention Leveraging Cacheable OperatioNs (M-FALCON)}
\label{sec:app-m-falcon}

\rebuttalrevision{In this section, we provide a detailed description of the M-FALCON algorithm discussed in~\cref{sec:design_e2e_inference}. We give pseudocode for M-FALCON in~\cref{alg:m-falcon}. M-FALCON introduces three key ideas.}

\begin{figure}[h]
    \begin{center}
    \includegraphics[width=0.8\linewidth]{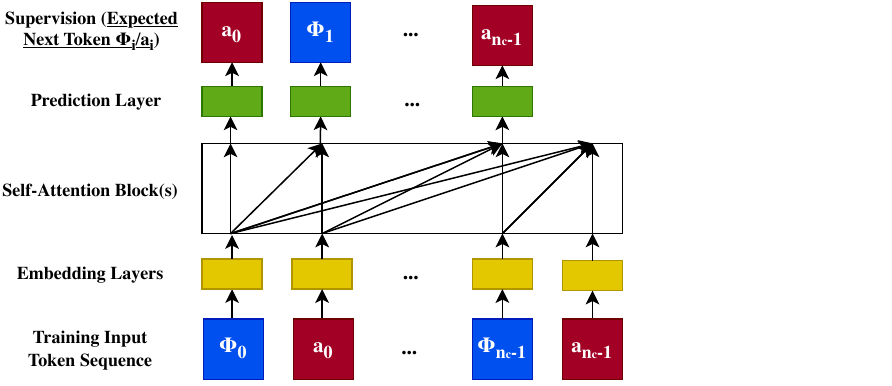}
    \subcaption{GR's ranking model training (with $n=2n_c$ tokens), in causal autoregressive settings.}
    \vspace{.5em}
    \includegraphics[width=0.8\linewidth]{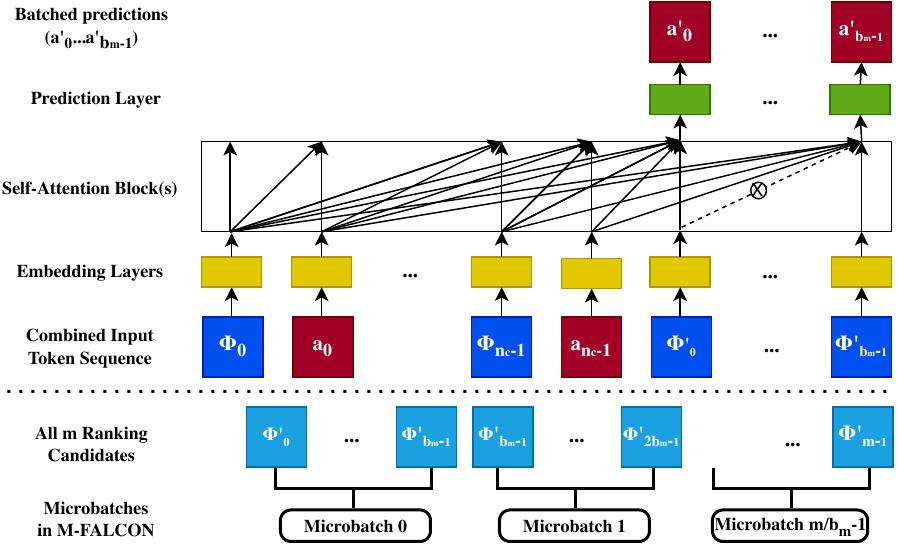}
    \subcaption{GR's ranking model inference utilizing the M-FALCON algorithm.}
    \end{center}
    \vspace{-1em}
    \caption{\rebuttalrevision{Illustration of the M-FALCON algorithm. Top: model training in GR's target-aware formulation. Bottom: model inference with $m$ candidates $\Phi'_0, \ldots, \Phi'_{m-1}$, divided into $\lceil m/b_m \rceil$ microbatches, where we show model inference for the first microbatch $\Phi'_0, \ldots, \Phi'_{b_m-1}$ (with $2n_c + b_m$ total tokens after $\Phi_0, a_0, \ldots, \Phi_{n_c-1}, a_{n_c-1}$ are taken into account) above the dotted line. Note that the self-attention algorithm is modified such that $\Phi'_i$ cannot attend to $\Phi'_j$ when $i \neq j$ -- this is highlighted with ``$\times$'' in the figure.}}
    \label{fig:app_mfalcon_training_inference}
    \vspace{-0.5em}
\end{figure}

\rebuttalrevision{\textbf{Batched inference can be applied to causal autoregressive settings.} The ranking task in GR is formulated in a target aware fashion as discussed~\cref{sec:dlrms-to-grs-tasks}. Common wisdom suggests that in a target-aware setting, we need to perform inference for one item at a time, with a cost of $O(m n^2d)$ for $m$ candidates and a sequence length of $n$. Here we show that this is not the optimal solution; even with vanilla Transformers, we can modify the attention mask used in self-attention to batch such operations (``batched inference'') and reduce cost to $O((n + m)^2d) = O(n^2d)$.}

\rebuttalrevision{An illustration is provided in~\cref{fig:app_mfalcon_training_inference}. Here, both \cref{fig:app_mfalcon_training_inference} (a) and (b) involve an attention mask matrix for causal autoregressive settings. The key difference is that \cref{fig:app_mfalcon_training_inference} (a) uses a standard lower triangular matrix of size $2n_c$ for causal training, whereas \cref{fig:app_mfalcon_training_inference} (b) modifies a lower triangular matrix of size $2n_c + b_m$ by setting entries for $(i, j)$s where $i, j \geq 2n_c, i \neq j$ to False or $-\infty$ to prevent target positions $\Phi'_0, \ldots, \Phi'_{b_m-1}$ from attending to each other. It is easy to see that by doing so, the output of the self-attention block for $\Phi'_i$, $a'_i$, only depends on $\Phi_0, a_0, \ldots, \Phi_{n_c-1}, a_{n_c-1}$, but not on $\Phi'_j$ ($i \neq j$). In other words, by making a forward pass over $(2n_c+b_m)$ tokens using the modified attention mask, we can now obtain the same results for the last $b_m$ tokens as if we've made $b_m$ separate forward passes over $(2n_c+1)$ tokens, with $\Phi'_i$ placed at the $2n_c$-th (0-based) position during the $i$-th forward pass utilizing a standard causal attention mask.}

\rebuttalrevision{\textbf{Microbatching scales batched inference to large candidate sets.} Ranking stage may need to deal with a large number of ranking candidates, up to tens of thousands~\citep{wang2020cold_preranking}. We can divide the overall $m$ candidates into $\lceil m / b_m\rceil$ microbatches of size $b_m$ such that $O(b_m) = O(n)$, which retains the $O((n + m)^2d) = O(n^2d)$ running time previously discussed for most practical recommender settings, up to tens of thousands of candidates.}

\rebuttalrevision{\textbf{Encoder-level caching enables compute sharing within and \textit{across} requests.} Finally, KV caching~\citep{pope2022scalinginference} can be applied both within and across requests. For instance, for the HSTU model presented in this work (\cref{sec:design_hstu_encoder}), $K(X)$ and $V(X)$ are fully cachable across microbatches within and/or across requests. For a cached forward pass, we only need to compute $U(X)$, $Q(X)$, $K(X)$, and $V(X)$ for the last $b_m$ tokens, while reusing cached $K(X)$ and $V(X)$ for the sequentialized user history containing $n$ tokens. $f_2(\text{Norm}{(A(X)V(X))} \odot U(X))$ similarly only needs to be recomputed for the $b_m$ candidates. This reduces the cached forward pass's computational complexity to $O(b_m d^2 + b_m nd)$, which significantly improves upon $O((n+b_m)d^2 + (n+b_m)^2d)$ by a factor of 2-4 even when $b_m=n$.}

\begin{algorithm}
\caption{\rebuttalrevision{M-FALCON Algorithm.}}
\label{alg:m-falcon}
\begin{algorithmic}[1]
    \STATE {\bfseries Input:} Merged token series $x_0, x_1, \ldots, x_{n-1}$ (can be e.g., $(\Phi_0, a_0, \ldots, \Phi_{n_c-1}, a_{n_c-1})$ where $n=2n_c$); {$m$ ranking candidates $\Phi'_0, \ldots, \Phi'_{m-1}$}; {a $b$-layer $h$-heads self-attention model trained in causal autoregressive settings (e.g., HSTU or Transformers)} $f(X, cacheStates, attnMask) \rightarrow (X', updatedCacheStates)$ where $X, X' \in \mathbb{R}^{N \times d}$, $attnMask \in \mathbb{R}^{N \times N}$, and $cachedStates, updatedCacheStates \in \mathbb{R}^{b \times h \times N \times d_{qk}} \times \mathbb{R}^{b \times h \times N \times d_{qk}}$ (due to caching $K(X)$s and $V(X)$s across $b$ layers); {microbatch size} $b_m$, where we assume $m$ is a multiple of $b_m$ for simplicity.
    \STATE {\bfseries Output: } Predictions for all $m$ ranking candidates, $(a'_0, \ldots, a'_{m-1})$.
    \STATE $numMicrobatches = (m + b_m - 1) // b_m$
    \STATE $attnMask = L_{n + b_m}$  \COMMENT{$L_{n+b_m}$ represents a lower triangular matrix. Lower triangular entries are 0s, the rest are $-\infty$.}
    \STATE $attnMask[i, j] = -\infty$ for $i, j \geq n, i \neq j$ \COMMENT{This prevents the last $b_m$ entries from attending to each other.}
    \STATE $(a'_0, a'_1, \ldots, a'_{b_m-1}), kvCache \leftarrow f(embLayer((x_0, x_1, \ldots, x_{n-1}, \Phi'_0, \ldots, \Phi'_{b_m-1})), \varnothing, attnMask)$
    \STATE $predictions = (a'_0, a'_1, \ldots, a'_{b_m-1})$ 
    \STATE $i = 1$
    \WHILE{$i < numMicrobatches$}
        \STATE $(a'_{b_m i}, a'_{b_m i + 1}, a'_{b_m (i + 1) -1}), \_ \leftarrow f(embLayer((x_0, x_1, \ldots, x_{n-1}, \Phi'_{b_m i}, \ldots, \Phi'_{b_m (i+1) - 1})), kvCache, attnMask)$
        \STATE $predictions \leftarrow predictions + (a'_{b_m i}, a'_{b_m i + 1}, \ldots, a'_{b_m (i + 1) -1})$
        \STATE $i \leftarrow i + 1$
    \ENDWHILE
    \STATE \textbf{return} $predictions$
\end{algorithmic}
\end{algorithm}

\rebuttalrevision{\cref{alg:m-falcon} is illustrated in~\cref{fig:app_mfalcon_training_inference} to help with understanding. We remark that M-FALCON is not only applicable to HSTUs and GRs, but also broadly applicable as an inference optimization algorithm for other target-aware causal autoregressive settings based on self-attention architectures.}

\subsection{Evaluation of Inference Throughput: Generative Recommenders (GRs) w/ M-FALCON vs DLRMs}
\label{sec:app-comparison-inference-throughput}

\rebuttalrevision{As discussed in \cref{sec:design_e2e_inference}, M-FALCON handles $b_m$ candidates in parallel to amortize computation costs across all $m$ candidates at inference time. To understand our design, we compare the throughput (i.e., the number of candidates scored per second, QPS) of GRs and DLRMs based on the same hardware setups.}

\begin{figure}[h]
    \vspace{-.7em}
    \begin{center}
\includegraphics[width=0.6\linewidth]{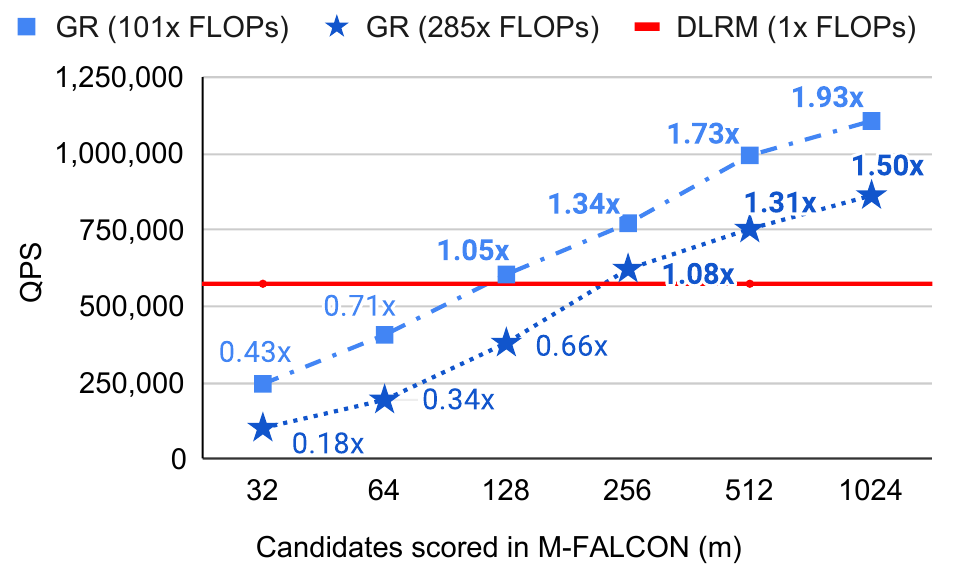}
    \end{center}
    \vspace{-1.3em}
    \caption{\rebuttalrevision{End-to-end inference throughtput: DLRMs vs GRs (w/ M-FALCON) in large-scale industrial settings. Note that this figure is the same as \cref{fig:gr-inference-throughput-v2}, and is reproduced here to facilitate reading.}}
    \label{fig:app_exp_scalability}
    \vspace{-1em}
\end{figure}

\begin{figure}[h]
    \vspace{-.7em}
    \begin{center}
    \includegraphics[width=0.8\linewidth]{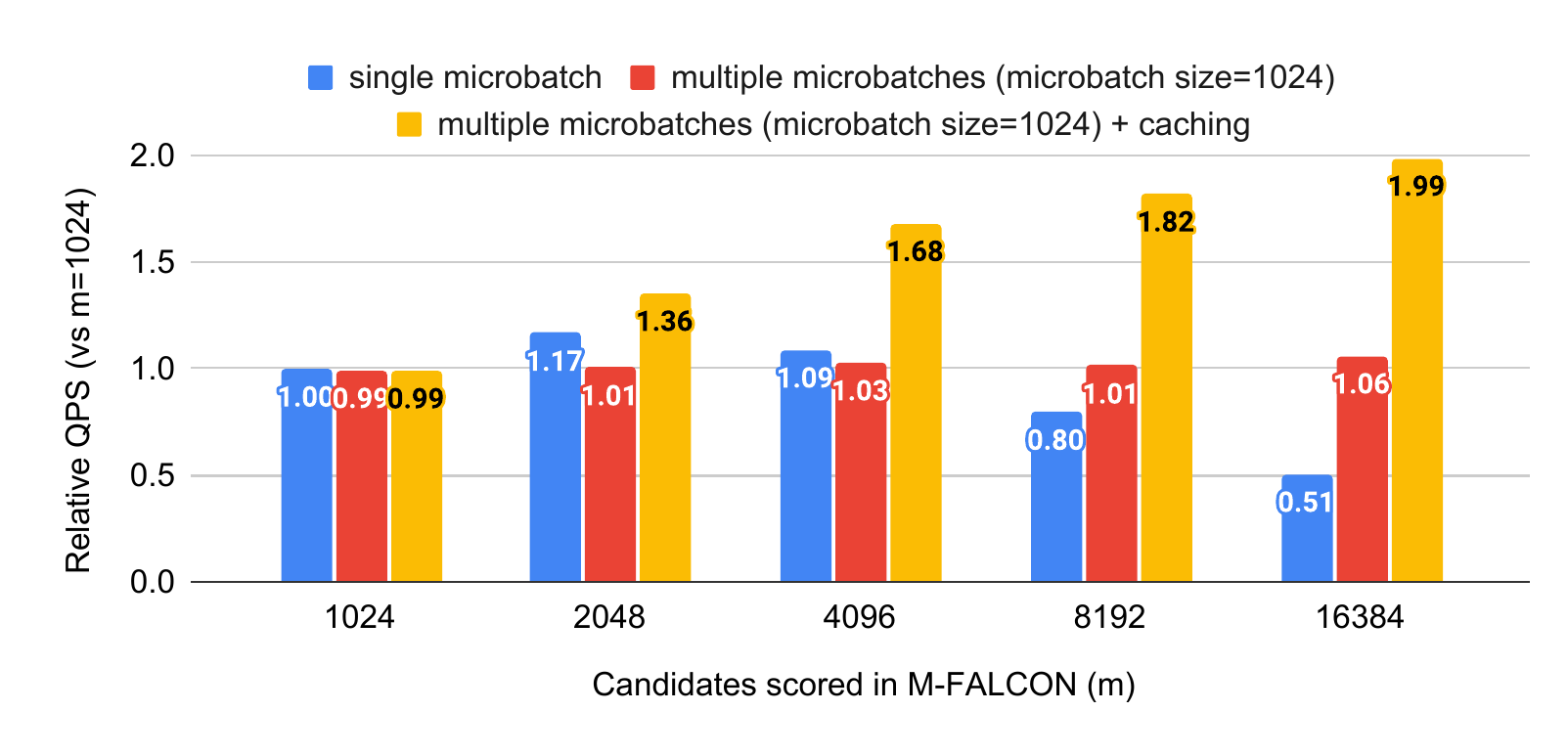}
    \end{center}
    \vspace{-2.5em}
    \caption{\rebuttalrevision{End-to-end inference throughtput: M-FALCON throughput scaling, on top of the 285x FLOPs GR model, in large batch settings where $m$ (total number of ranking candidates) ranges from 1024 to 16384, and $b_m = 1024$.}}
    \label{fig:app_exp_scalability_1024plus}
    \vspace{-1em}
\end{figure}

\rebuttalrevision{As shown in~\cref{fig:app_exp_scalability} and~\cref{fig:app_exp_scalability_1024plus}, GRs' throughput scales in a sublinear way based on the number of ranking-stage candidates ($m$), up to a certain region -- $m=2048$ in our case study -- due to \textit{batched inference} enabling cost amortization. This confirms the criticality of batched inference in causal autoregressive settings. 
Due to attention complexity scaling as $O((n + b_m)^2)$, leveraging multiple microbatches by itself improves throughput. Caching further eliminates redundant linear and attention computations on top of microbatching. The two combined resulted in up to 1.99x additional speedups relative to the $b_m = m = 1024$ baseline using a single microbatch, as shown in~\cref{fig:app_exp_scalability_1024plus}. Overall, with the efficient HSTU encoder design and utilizing M-FALCON, HSTU-based Generative Recommenders \emph{outperform} DLRMs in terms of throughput on a large-scale production setup by up to 2.99x, despite GRs being \emph{285x more complex in terms of FLOPs}.}

%% file: main.bbl
\begin{thebibliography}{70}
\providecommand{\natexlab}[1]{#1}
\providecommand{\url}[1]{\texttt{#1}}
\expandafter\ifx\csname urlstyle\endcsname\relax
  \providecommand{\doi}[1]{doi: #1}\else
  \providecommand{\doi}{doi: \begingroup \urlstyle{rm}\Url}\fi

\bibitem[Bao et~al.(2023)Bao, Zhang, Zhang, Wang, Feng, and He]{Bao_2023}
Bao, K., Zhang, J., Zhang, Y., Wang, W., Feng, F., and He, X.
\newblock Tallrec: An effective and efficient tuning framework to align large language model with recommendation.
\newblock In \emph{Proceedings of the 17th ACM Conference on Recommender Systems}, RecSys ’23. ACM, September 2023.
\newblock \doi{10.1145/3604915.3608857}.
\newblock URL \url{http://dx.doi.org/10.1145/3604915.3608857}.

\bibitem[Brown et~al.(2020)Brown, Mann, Ryder, Subbiah, Kaplan, Dhariwal, Neelakantan, Shyam, Sastry, Askell, Agarwal, Herbert-Voss, Krueger, Henighan, Child, Ramesh, Ziegler, Wu, Winter, Hesse, Chen, Sigler, Litwin, Gray, Chess, Clark, Berner, McCandlish, Radford, Sutskever, and Amodei]{brown2020language}
Brown, T.~B., Mann, B., Ryder, N., Subbiah, M., Kaplan, J., Dhariwal, P., Neelakantan, A., Shyam, P., Sastry, G., Askell, A., Agarwal, S., Herbert-Voss, A., Krueger, G., Henighan, T., Child, R., Ramesh, A., Ziegler, D.~M., Wu, J., Winter, C., Hesse, C., Chen, M., Sigler, E., Litwin, M., Gray, S., Chess, B., Clark, J., Berner, C., McCandlish, S., Radford, A., Sutskever, I., and Amodei, D.
\newblock Language models are few-shot learners.
\newblock 2020.

\bibitem[Chang et~al.(2023)Chang, Zhang, Fu, Zang, Guan, Lu, Hui, Leng, Niu, Song, and Gai]{kwai2023twin}
Chang, J., Zhang, C., Fu, Z., Zang, X., Guan, L., Lu, J., Hui, Y., Leng, D., Niu, Y., Song, Y., and Gai, K.
\newblock Twin: Two-stage interest network for lifelong user behavior modeling in ctr prediction at kuaishou, 2023.

\bibitem[Chen et~al.(2019)Chen, Zhao, Li, Huang, and Ou]{bst_dlpkdd19}
Chen, Q., Zhao, H., Li, W., Huang, P., and Ou, W.
\newblock Behavior sequence transformer for e-commerce recommendation in alibaba.
\newblock In \emph{Proceedings of the 1st International Workshop on Deep Learning Practice for High-Dimensional Sparse Data}, DLP-KDD '19, New York, NY, USA, 2019. Association for Computing Machinery.
\newblock ISBN 9781450367837.
\newblock \doi{10.1145/3326937.3341261}.
\newblock URL \url{https://doi.org/10.1145/3326937.3341261}.

\bibitem[Chen et~al.(2020)Chen, Kornblith, Norouzi, and Hinton]{simclr_icml20}
Chen, T., Kornblith, S., Norouzi, M., and Hinton, G.
\newblock A simple framework for contrastive learning of visual representations.
\newblock In \emph{Proceedings of the 37th International Conference on Machine Learning}, ICML'20, 2020.

\bibitem[Cheng et~al.(2016)Cheng, Koc, Harmsen, Shaked, Chandra, Aradhye, Anderson, Corrado, Chai, Ispir, Anil, Haque, Hong, Jain, Liu, and Shah]{wdl_goog_dlrs16}
Cheng, H.-T., Koc, L., Harmsen, J., Shaked, T., Chandra, T., Aradhye, H., Anderson, G., Corrado, G., Chai, W., Ispir, M., Anil, R., Haque, Z., Hong, L., Jain, V., Liu, X., and Shah, H.
\newblock Wide \& deep learning for recommender systems.
\newblock In \emph{Proceedings of the 1st Workshop on Deep Learning for Recommender Systems}, DLRS 2016, pp.\  7–10, 2016.
\newblock ISBN 9781450347952.

\bibitem[Child et~al.(2019)Child, Gray, Radford, and Sutskever]{sparsetransformers_openai19}
Child, R., Gray, S., Radford, A., and Sutskever, I.
\newblock Generating long sequences with sparse transformers.
\newblock \emph{CoRR}, abs/1904.10509, 2019.
\newblock URL \url{http://arxiv.org/abs/1904.10509}.

\bibitem[Covington et~al.(2016)Covington, Adams, and Sargin]{ytdnn_goog_recsys16}
Covington, P., Adams, J., and Sargin, E.
\newblock Deep neural networks for youtube recommendations.
\newblock In \emph{Proceedings of the 10th ACM Conference on Recommender Systems}, RecSys '16, pp.\  191–198, 2016.
\newblock ISBN 9781450340359.

\bibitem[Cui et~al.(2022)Cui, Ma, Zhou, Zhou, and Yang]{cui2022m6rec}
Cui, Z., Ma, J., Zhou, C., Zhou, J., and Yang, H.
\newblock M6-rec: Generative pretrained language models are open-ended recommender systems, 2022.

\bibitem[Dallmann et~al.(2021)Dallmann, Zoller, and Hotho]{seqreceval_recsys21}
Dallmann, A., Zoller, D., and Hotho, A.
\newblock A case study on sampling strategies for evaluating neural sequential item recommendation models.
\newblock In \emph{Proceedings of the 15th ACM Conference on Recommender Systems}, RecSys '21, pp.\  505–514, 2021.
\newblock ISBN 9781450384582.

\bibitem[Dao(2023)]{dao2023flashattention2}
Dao, T.
\newblock Flashattention-2: Faster attention with better parallelism and work partitioning, 2023.

\bibitem[Dao et~al.(2022)Dao, Fu, Ermon, Rudra, and R{\'e}]{dao2022flashattention}
Dao, T., Fu, D.~Y., Ermon, S., Rudra, A., and R{\'e}, C.
\newblock Flash{A}ttention: Fast and memory-efficient exact attention with {IO}-awareness.
\newblock In \emph{Advances in Neural Information Processing Systems}, 2022.

\bibitem[Devlin et~al.(2019)Devlin, Chang, Lee, and Toutanova]{bert_naacl19}
Devlin, J., Chang, M., Lee, K., and Toutanova, K.
\newblock {BERT:} pre-training of deep bidirectional transformers for language understanding.
\newblock In Burstein, J., Doran, C., and Solorio, T. (eds.), \emph{Proceedings of the 2019 Conference of the North American Chapter of the Association for Computational Linguistics: Human Language Technologies, {NAACL-HLT} 2019, Minneapolis, MN, USA, June 2-7, 2019, Volume 1 (Long and Short Papers)}, pp.\  4171--4186. Association for Computational Linguistics, 2019.
\newblock \doi{10.18653/v1/n19-1423}.
\newblock URL \url{https://doi.org/10.18653/v1/n19-1423}.

\bibitem[Eksombatchai et~al.(2018)Eksombatchai, Jindal, Liu, Liu, Sharma, Sugnet, Ulrich, and Leskovec]{pixie_pins_www18}
Eksombatchai, C., Jindal, P., Liu, J.~Z., Liu, Y., Sharma, R., Sugnet, C., Ulrich, M., and Leskovec, J.
\newblock Pixie: A system for recommending 3+ billion items to 200+ million users in real-time.
\newblock In \emph{Proceedings of the 2018 World Wide Web Conference}, WWW '18, pp.\  1775–1784, 2018.
\newblock ISBN 9781450356398.

\bibitem[Elfwing et~al.(2017)Elfwing, Uchibe, and Doya]{elfwing2017silu}
Elfwing, S., Uchibe, E., and Doya, K.
\newblock Sigmoid-weighted linear units for neural network function approximation in reinforcement learning.
\newblock \emph{CoRR}, abs/1702.03118, 2017.
\newblock URL \url{http://arxiv.org/abs/1702.03118}.

\bibitem[Gao et~al.(2021)Gao, Fan, Wang, Sun, Jia, Xiao, Ding, Bin, Yang, and Liu]{dr_cikm21}
Gao, W., Fan, X., Wang, C., Sun, J., Jia, K., Xiao, W., Ding, R., Bin, X., Yang, H., and Liu, X.
\newblock Learning an end-to-end structure for retrieval in large-scale recommendations.
\newblock In \emph{Proceedings of the 30th ACM International Conference on Information and Knowledge Management}, CIKM '21, pp.\  524–533, 2021.
\newblock ISBN 9781450384469.

\bibitem[Gillenwater et~al.(2014)Gillenwater, Kulesza, Fox, and Taskar]{dpp_neurips14}
Gillenwater, J., Kulesza, A., Fox, E., and Taskar, B.
\newblock Expectation-maximization for learning determinantal point processes.
\newblock In \emph{Proceedings of the 27th International Conference on Neural Information Processing Systems - Volume 2}, NIPS'14, pp.\  3149–3157, Cambridge, MA, USA, 2014. MIT Press.

\bibitem[Gu et~al.(2022)Gu, Goel, and R{\'{e}}]{s4_iclr22}
Gu, A., Goel, K., and R{\'{e}}, C.
\newblock Efficiently modeling long sequences with structured state spaces.
\newblock In \emph{The Tenth International Conference on Learning Representations, {ICLR} 2022, Virtual Event, April 25-29, 2022}. OpenReview.net, 2022.
\newblock URL \url{https://openreview.net/forum?id=uYLFoz1vlAC}.

\bibitem[Guo et~al.(2017)Guo, Tang, Ye, Li, and He]{deepfm_ijcai17}
Guo, H., Tang, R., Ye, Y., Li, Z., and He, X.
\newblock Deepfm: A factorization-machine based neural network for ctr prediction.
\newblock In \emph{Proceedings of the 26th International Joint Conference on Artificial Intelligence}, IJCAI'17, pp.\  1725–1731, 2017.
\newblock ISBN 9780999241103.

\bibitem[Gupta et~al.(2014)Gupta, Bengio, and Weston]{rowwiseadagrad_jmlr14}
Gupta, M.~R., Bengio, S., and Weston, J.
\newblock Training highly multiclass classifiers.
\newblock \emph{J. Mach. Learn. Res.}, 15\penalty0 (1):\penalty0 1461–1492, jan 2014.
\newblock ISSN 1532-4435.

\bibitem[He et~al.(2015)He, Zhang, Ren, and Sun]{He2015_resnet}
He, K., Zhang, X., Ren, S., and Sun, J.
\newblock Deep residual learning for image recognition.
\newblock \emph{arXiv preprint arXiv:1512.03385}, 2015.

\bibitem[He et~al.(2014)He, Pan, Jin, Xu, Liu, Xu, Shi, Atallah, Herbrich, Bowers, and Candela]{metaads}
He, X., Pan, J., Jin, O., Xu, T., Liu, B., Xu, T., Shi, Y., Atallah, A., Herbrich, R., Bowers, S., and Candela, J.~Q.
\newblock Practical lessons from predicting clicks on ads at facebook.
\newblock In \emph{ADKDD'14: Proceedings of the Eighth International Workshop on Data Mining for Online Advertising}, New York, NY, USA, 2014. Association for Computing Machinery.
\newblock ISBN 9781450329996.

\bibitem[Hidasi et~al.(2016)Hidasi, Karatzoglou, Baltrunas, and Tikk]{gru4rec_iclr16}
Hidasi, B., Karatzoglou, A., Baltrunas, L., and Tikk, D.
\newblock Session-based recommendations with recurrent neural networks.
\newblock In Bengio, Y. and LeCun, Y. (eds.), \emph{4th International Conference on Learning Representations, {ICLR} 2016, San Juan, Puerto Rico, May 2-4, 2016, Conference Track Proceedings}, 2016.
\newblock URL \url{http://arxiv.org/abs/1511.06939}.

\bibitem[Hou et~al.(2024)Hou, Zhang, Lin, Lu, Xie, McAuley, and Zhao]{hou2024large}
Hou, Y., Zhang, J., Lin, Z., Lu, H., Xie, R., McAuley, J., and Zhao, W.~X.
\newblock Large language models are zero-shot rankers for recommender systems.
\newblock In \emph{Advances in Information Retrieval - 46th European Conference on {IR} Research, {ECIR} 2024}, 2024.

\bibitem[Hua et~al.(2022)Hua, Dai, Liu, and Le]{icml2022flash}
Hua, W., Dai, Z., Liu, H., and Le, Q.~V.
\newblock Transformer quality in linear time.
\newblock In Chaudhuri, K., Jegelka, S., Song, L., Szepesv{\'{a}}ri, C., Niu, G., and Sabato, S. (eds.), \emph{International Conference on Machine Learning, {ICML} 2022, 17-23 July 2022, Baltimore, Maryland, {USA}}, volume 162 of \emph{Proceedings of Machine Learning Research}, pp.\  9099--9117. {PMLR}, 2022.
\newblock URL \url{https://proceedings.mlr.press/v162/hua22a.html}.

\bibitem[Huang et~al.(2016)Huang, Sun, Liu, Sedra, and Weinberger]{huang2016stochasticdepth}
Huang, G., Sun, Y., Liu, Z., Sedra, D., and Weinberger, K.
\newblock Deep networks with stochastic depth, 2016.

\bibitem[Jegou et~al.(2011)Jegou, Douze, and Schmid]{pq_nns_pami11}
Jegou, H., Douze, M., and Schmid, C.
\newblock Product quantization for nearest neighbor search.
\newblock \emph{IEEE Trans. Pattern Anal. Mach. Intell.}, 33\penalty0 (1):\penalty0 117–128, jan 2011.
\newblock ISSN 0162-8828.
\newblock \doi{10.1109/TPAMI.2010.57}.
\newblock URL \url{https://doi.org/10.1109/TPAMI.2010.57}.

\bibitem[Kang \& McAuley(2018)Kang and McAuley]{sasrec_icdm18}
Kang, W.-C. and McAuley, J.
\newblock Self-attentive sequential recommendation.
\newblock In \emph{2018 International Conference on Data Mining (ICDM)}, pp.\  197--206, 2018.

\bibitem[Kaplan et~al.(2020)Kaplan, McCandlish, Henighan, Brown, Chess, Child, Gray, Radford, Wu, and Amodei]{kaplan2020scalinglaws}
Kaplan, J., McCandlish, S., Henighan, T., Brown, T.~B., Chess, B., Child, R., Gray, S., Radford, A., Wu, J., and Amodei, D.
\newblock Scaling laws for neural language models.
\newblock \emph{CoRR}, abs/2001.08361, 2020.
\newblock URL \url{https://arxiv.org/abs/2001.08361}.

\bibitem[Katharopoulos et~al.(2020)Katharopoulos, Vyas, Pappas, and Fleuret]{linearattention_icml20}
Katharopoulos, A., Vyas, A., Pappas, N., and Fleuret, F.
\newblock Transformers are rnns: Fast autoregressive transformers with linear attention.
\newblock In \emph{Proceedings of the 37th International Conference on Machine Learning}, ICML'20. JMLR.org, 2020.

\bibitem[Khudia et~al.(2021)Khudia, Huang, Basu, Deng, Liu, Park, and Smelyanskiy]{fbgemm21arxiv}
Khudia, D., Huang, J., Basu, P., Deng, S., Liu, H., Park, J., and Smelyanskiy, M.
\newblock Fbgemm: Enabling high-performance low-precision deep learning inference.
\newblock \emph{arXiv preprint arXiv:2101.05615}, 2021.

\bibitem[Klenitskiy \& Vasilev(2023)Klenitskiy and Vasilev]{bert4recvssasrec_recsys23}
Klenitskiy, A. and Vasilev, A.
\newblock Turning dross into gold loss: is bert4rec really better than sasrec?
\newblock In \emph{Proceedings of the 17th ACM Conference on Recommender Systems}, RecSys '23, pp.\  1120–1125, New York, NY, USA, 2023. Association for Computing Machinery.
\newblock ISBN 9798400702419.
\newblock \doi{10.1145/3604915.3610644}.
\newblock URL \url{https://doi.org/10.1145/3604915.3610644}.

\bibitem[Korthikanti et~al.(2022)Korthikanti, Casper, Lym, McAfee, Andersch, Shoeybi, and Catanzaro]{korthikanti2022reducing}
Korthikanti, V., Casper, J., Lym, S., McAfee, L., Andersch, M., Shoeybi, M., and Catanzaro, B.
\newblock Reducing activation recomputation in large transformer models, 2022.

\bibitem[Li et~al.(2002)Li, Chang, Garcia-Molina, and Wiederhold]{clusteringss_tkde02}
Li, C., Chang, E., Garcia-Molina, H., and Wiederhold, G.
\newblock Clustering for approximate similarity search in high-dimensional spaces.
\newblock \emph{IEEE Transactions on Knowledge and Data Engineering}, 14\penalty0 (4):\penalty0 792--808, 2002.

\bibitem[Li et~al.(2023)Li, Wang, Li, Fu, Shen, Shang, and McAuley]{li23text}
Li, J., Wang, M., Li, J., Fu, J., Shen, X., Shang, J., and McAuley, J.
\newblock Text is all you need: Learning language representations for sequential recommendation.
\newblock In \emph{KDD}, 2023.

\bibitem[Liu et~al.(2022)Liu, Zou, Zou, Wang, Zhang, Tang, Zhu, Zhu, Wu, Wang, and Cheng]{bd2022monolith}
Liu, Z., Zou, L., Zou, X., Wang, C., Zhang, B., Tang, D., Zhu, B., Zhu, Y., Wu, P., Wang, K., and Cheng, Y.
\newblock Monolith: Real time recommendation system with collisionless embedding table, 2022.

\bibitem[Ma et~al.(2018)Ma, Zhao, Yi, Chen, Hong, and Chi]{mmoe_kdd18}
Ma, J., Zhao, Z., Yi, X., Chen, J., Hong, L., and Chi, E.~H.
\newblock Modeling task relationships in multi-task learning with multi-gate mixture-of-experts.
\newblock KDD '18, 2018.

\bibitem[Mudigere et~al.(2022)Mudigere, Hao, Huang, Jia, Tulloch, Sridharan, Liu, Ozdal, Nie, Park, Luo, Yang, Gao, Ivchenko, Basant, Hu, Yang, Ardestani, Wang, Komuravelli, Chu, Yilmaz, Li, Qian, Feng, Ma, Yang, Wen, Li, Yang, Sun, Zhao, Melts, Dhulipala, Kishore, Graf, Eisenman, Matam, Gangidi, Chen, Krishnan, Nayak, Nair, Muthiah, khorashadi, Bhattacharya, Lapukhov, Naumov, Mathews, Qiao, Smelyanskiy, Jia, and Rao]{dlrm_isca22}
Mudigere, D., Hao, Y., Huang, J., Jia, Z., Tulloch, A., Sridharan, S., Liu, X., Ozdal, M., Nie, J., Park, J., Luo, L., Yang, J.~A., Gao, L., Ivchenko, D., Basant, A., Hu, Y., Yang, J., Ardestani, E.~K., Wang, X., Komuravelli, R., Chu, C.-H., Yilmaz, S., Li, H., Qian, J., Feng, Z., Ma, Y., Yang, J., Wen, E., Li, H., Yang, L., Sun, C., Zhao, W., Melts, D., Dhulipala, K., Kishore, K., Graf, T., Eisenman, A., Matam, K.~K., Gangidi, A., Chen, G.~J., Krishnan, M., Nayak, A., Nair, K., Muthiah, B., khorashadi, M., Bhattacharya, P., Lapukhov, P., Naumov, M., Mathews, A., Qiao, L., Smelyanskiy, M., Jia, B., and Rao, V.
\newblock Software-hardware co-design for fast and scalable training of deep learning recommendation models.
\newblock In \emph{Proceedings of the 49th Annual International Symposium on Computer Architecture}, ISCA '22, pp.\  993–1011, New York, NY, USA, 2022. Association for Computing Machinery.
\newblock ISBN 9781450386104.
\newblock \doi{10.1145/3470496.3533727}.
\newblock URL \url{https://doi.org/10.1145/3470496.3533727}.

\bibitem[Peng et~al.(2024)Peng, Quesnelle, Fan, and Shippole]{peng2024yarn}
Peng, B., Quesnelle, J., Fan, H., and Shippole, E.
\newblock Ya{RN}: Efficient context window extension of large language models.
\newblock In \emph{The Twelfth International Conference on Learning Representations}, 2024.
\newblock URL \url{https://openreview.net/forum?id=wHBfxhZu1u}.

\bibitem[Pope et~al.(2022)Pope, Douglas, Chowdhery, Devlin, Bradbury, Levskaya, Heek, Xiao, Agrawal, and Dean]{pope2022scalinginference}
Pope, R., Douglas, S., Chowdhery, A., Devlin, J., Bradbury, J., Levskaya, A., Heek, J., Xiao, K., Agrawal, S., and Dean, J.
\newblock Efficiently scaling transformer inference, 2022.

\bibitem[Press et~al.(2022)Press, Smith, and Lewis]{lengthextrapolation2022iclr}
Press, O., Smith, N.~A., and Lewis, M.
\newblock Train short, test long: Attention with linear biases enables input length extrapolation.
\newblock In \emph{The Tenth International Conference on Learning Representations, {ICLR} 2022, Virtual Event, April 25-29, 2022}. OpenReview.net, 2022.
\newblock URL \url{https://openreview.net/forum?id=R8sQPpGCv0}.

\bibitem[Rabe \& Staats(2021)Rabe and Staats]{rabe2021memoryefficientselfattention}
Rabe, M.~N. and Staats, C.
\newblock Self-attention does not need $o(n^2)$ memory, 2021.

\bibitem[Raffel et~al.(2020)Raffel, Shazeer, Roberts, Lee, Narang, Matena, Zhou, Li, and Liu]{t5_jmlr20}
Raffel, C., Shazeer, N., Roberts, A., Lee, K., Narang, S., Matena, M., Zhou, Y., Li, W., and Liu, P.~J.
\newblock Exploring the limits of transfer learning with a unified text-to-text transformer.
\newblock \emph{J. Mach. Learn. Res.}, 21\penalty0 (1), jan 2020.
\newblock ISSN 1532-4435.

\bibitem[Rendle(2010)]{fm_rendle_icdm10}
Rendle, S.
\newblock Factorization machines.
\newblock In \emph{2010 IEEE International Conference on Data Mining (ICDM)}, pp.\  995--1000, 2010.
\newblock \doi{10.1109/ICDM.2010.127}.

\bibitem[Rendle et~al.(2020)Rendle, Krichene, Zhang, and Anderson]{ncf_mf_goog_recsys20}
Rendle, S., Krichene, W., Zhang, L., and Anderson, J.
\newblock Neural collaborative filtering vs. matrix factorization revisited.
\newblock In \emph{Fourteenth ACM Conference on Recommender Systems (RecSys'20)}, pp.\  240–248, 2020.
\newblock ISBN 9781450375832.

\bibitem[Shazeer(2020)]{shazeer2020glu}
Shazeer, N.
\newblock Glu variants improve transformer, 2020.

\bibitem[Shin et~al.(2023)Shin, Kwak, Kim, Ramstr\"{o}m, Jeong, Ha, and Kim]{scalingreco2023aaai}
Shin, K., Kwak, H., Kim, S.~Y., Ramstr\"{o}m, M.~N., Jeong, J., Ha, J.-W., and Kim, K.-M.
\newblock Scaling law for recommendation models: towards general-purpose user representations.
\newblock In \emph{Proceedings of the Thirty-Seventh AAAI Conference on Artificial Intelligence and Thirty-Fifth Conference on Innovative Applications of Artificial Intelligence and Thirteenth Symposium on Educational Advances in Artificial Intelligence}, AAAI'23/IAAI'23/EAAI'23. AAAI Press, 2023.
\newblock ISBN 978-1-57735-880-0.
\newblock \doi{10.1609/aaai.v37i4.25582}.
\newblock URL \url{https://doi.org/10.1609/aaai.v37i4.25582}.

\bibitem[Shrivastava \& Li(2014)Shrivastava and Li]{alsh_ping_neurips2014}
Shrivastava, A. and Li, P.
\newblock Asymmetric lsh (alsh) for sublinear time maximum inner product search (mips).
\newblock In \emph{Advances in Neural Information Processing Systems}, volume~27, 2014.

\bibitem[Sileo et~al.(2022)Sileo, Vossen, and Raymaekers]{sileo_zeroshot_lmrec_ecir22}
Sileo, D., Vossen, W., and Raymaekers, R.
\newblock Zero-shot recommendation as language modeling.
\newblock In Hagen, M., Verberne, S., Macdonald, C., Seifert, C., Balog, K., N{\o}rv{\aa}g, K., and Setty, V. (eds.), \emph{Advances in Information Retrieval - 44th European Conference on {IR} Research, {ECIR} 2022, Stavanger, Norway, April 10-14, 2022, Proceedings, Part {II}}, volume 13186 of \emph{Lecture Notes in Computer Science}, pp.\  223--230. Springer, 2022.
\newblock \doi{10.1007/978-3-030-99739-7\_26}.
\newblock URL \url{https://doi.org/10.1007/978-3-030-99739-7\_26}.

\bibitem[Su et~al.(2023)Su, Lu, Pan, Murtadha, Wen, and Liu]{su2023roformer}
Su, J., Lu, Y., Pan, S., Murtadha, A., Wen, B., and Liu, Y.
\newblock Roformer: Enhanced transformer with rotary position embedding, 2023.

\bibitem[Sun et~al.(2019)Sun, Liu, Wu, Pei, Lin, Ou, and Jiang]{bert4rec_cikm19}
Sun, F., Liu, J., Wu, J., Pei, C., Lin, X., Ou, W., and Jiang, P.
\newblock Bert4rec: Sequential recommendation with bidirectional encoder representations from transformer.
\newblock In \emph{Proceedings of the 28th ACM International Conference on Information and Knowledge Management}, CIKM '19, pp.\  1441–1450, 2019.
\newblock ISBN 9781450369763.

\bibitem[Tang et~al.(2020)Tang, Liu, Zhao, and Gong]{ple_recsys20}
Tang, H., Liu, J., Zhao, M., and Gong, X.
\newblock Progressive layered extraction (ple): A novel multi-task learning (mtl) model for personalized recommendations.
\newblock In \emph{Proceedings of the 14th ACM Conference on Recommender Systems}, RecSys '20, pp.\  269–278, New York, NY, USA, 2020. Association for Computing Machinery.
\newblock ISBN 9781450375832.
\newblock \doi{10.1145/3383313.3412236}.
\newblock URL \url{https://doi.org/10.1145/3383313.3412236}.

\bibitem[Touvron et~al.(2023{\natexlab{a}})Touvron, Lavril, Izacard, Martinet, Lachaux, Lacroix, Rozière, Goyal, Hambro, Azhar, Rodriguez, Joulin, Grave, and Lample]{touvron2023llama1}
Touvron, H., Lavril, T., Izacard, G., Martinet, X., Lachaux, M.-A., Lacroix, T., Rozière, B., Goyal, N., Hambro, E., Azhar, F., Rodriguez, A., Joulin, A., Grave, E., and Lample, G.
\newblock Llama: Open and efficient foundation language models, 2023{\natexlab{a}}.

\bibitem[Touvron et~al.(2023{\natexlab{b}})Touvron, Martin, Stone, Albert, Almahairi, Babaei, Bashlykov, Batra, Bhargava, Bhosale, Bikel, Blecher, Ferrer, Chen, Cucurull, Esiobu, Fernandes, Fu, Fu, Fuller, Gao, Goswami, Goyal, Hartshorn, Hosseini, Hou, Inan, Kardas, Kerkez, Khabsa, Kloumann, Korenev, Koura, Lachaux, Lavril, Lee, Liskovich, Lu, Mao, Martinet, Mihaylov, Mishra, Molybog, Nie, Poulton, Reizenstein, Rungta, Saladi, Schelten, Silva, Smith, Subramanian, Tan, Tang, Taylor, Williams, Kuan, Xu, Yan, Zarov, Zhang, Fan, Kambadur, Narang, Rodriguez, Stojnic, Edunov, and Scialom]{touvron2023llama}
Touvron, H., Martin, L., Stone, K., Albert, P., Almahairi, A., Babaei, Y., Bashlykov, N., Batra, S., Bhargava, P., Bhosale, S., Bikel, D., Blecher, L., Ferrer, C.~C., Chen, M., Cucurull, G., Esiobu, D., Fernandes, J., Fu, J., Fu, W., Fuller, B., Gao, C., Goswami, V., Goyal, N., Hartshorn, A., Hosseini, S., Hou, R., Inan, H., Kardas, M., Kerkez, V., Khabsa, M., Kloumann, I., Korenev, A., Koura, P.~S., Lachaux, M.-A., Lavril, T., Lee, J., Liskovich, D., Lu, Y., Mao, Y., Martinet, X., Mihaylov, T., Mishra, P., Molybog, I., Nie, Y., Poulton, A., Reizenstein, J., Rungta, R., Saladi, K., Schelten, A., Silva, R., Smith, E.~M., Subramanian, R., Tan, X.~E., Tang, B., Taylor, R., Williams, A., Kuan, J.~X., Xu, P., Yan, Z., Zarov, I., Zhang, Y., Fan, A., Kambadur, M., Narang, S., Rodriguez, A., Stojnic, R., Edunov, S., and Scialom, T.
\newblock Llama 2: Open foundation and fine-tuned chat models, 2023{\natexlab{b}}.

\bibitem[Vaswani et~al.(2017)Vaswani, Shazeer, Parmar, Uszkoreit, Jones, Gomez, Kaiser, and Polosukhin]{transformers_goog_neurips17}
Vaswani, A., Shazeer, N., Parmar, N., Uszkoreit, J., Jones, L., Gomez, A.~N., Kaiser, L., and Polosukhin, I.
\newblock Attention is all you need.
\newblock In \emph{Proceedings of the 31st International Conference on Neural Information Processing Systems}, NIPS'17, pp.\  6000–6010, 2017.
\newblock ISBN 9781510860964.

\bibitem[Wang et~al.(2021)Wang, Shivanna, Cheng, Jain, Lin, Hong, and Chi]{dcnv2_www21}
Wang, R., Shivanna, R., Cheng, D., Jain, S., Lin, D., Hong, L., and Chi, E.
\newblock Dcn v2: Improved deep \& cross network and practical lessons for web-scale learning to rank systems.
\newblock In \emph{Proceedings of the Web Conference 2021}, WWW '21, pp.\  1785–1797, New York, NY, USA, 2021. Association for Computing Machinery.
\newblock ISBN 9781450383127.
\newblock \doi{10.1145/3442381.3450078}.
\newblock URL \url{https://doi.org/10.1145/3442381.3450078}.

\bibitem[Wang et~al.(2020)Wang, Zhao, Jiang, Zhou, Zhu, and Gai]{wang2020cold_preranking}
Wang, Z., Zhao, L., Jiang, B., Zhou, G., Zhu, X., and Gai, K.
\newblock Cold: Towards the next generation of pre-ranking system, 2020.

\bibitem[Xia et~al.(2023)Xia, Eksombatchai, Pancha, Badani, Wang, Gu, Joshi, Farahpour, Zhang, and Zhai]{pinterest23transact}
Xia, X., Eksombatchai, P., Pancha, N., Badani, D.~D., Wang, P.-W., Gu, N., Joshi, S.~V., Farahpour, N., Zhang, Z., and Zhai, A.
\newblock Transact: Transformer-based realtime user action model for recommendation at pinterest.
\newblock In \emph{Proceedings of the 29th ACM SIGKDD Conference on Knowledge Discovery and Data Mining}, KDD '23, pp.\  5249–5259, New York, NY, USA, 2023. Association for Computing Machinery.
\newblock ISBN 9798400701030.
\newblock \doi{10.1145/3580305.3599918}.
\newblock URL \url{https://doi.org/10.1145/3580305.3599918}.

\bibitem[Xiao et~al.(2017)Xiao, Ye, He, Zhang, Wu, and Chua]{afm_ijcai17}
Xiao, J., Ye, H., He, X., Zhang, H., Wu, F., and Chua, T.-S.
\newblock Attentional factorization machines: Learning the weight of feature interactions via attention networks.
\newblock In \emph{Proceedings of the 26th International Joint Conference on Artificial Intelligence}, IJCAI'17, pp.\  3119–3125. AAAI Press, 2017.
\newblock ISBN 9780999241103.

\bibitem[Xiong et~al.(2020)Xiong, Yang, He, Zheng, Zheng, Xing, Zhang, Lan, Wang, and Liu]{transformer_ln_icml20}
Xiong, R., Yang, Y., He, D., Zheng, K., Zheng, S., Xing, C., Zhang, H., Lan, Y., Wang, L., and Liu, T.-Y.
\newblock On layer normalization in the transformer architecture.
\newblock In \emph{Proceedings of the 37th International Conference on Machine Learning}, ICML'20. JMLR.org, 2020.

\bibitem[Yang et~al.(2020)Yang, Yi, Zhiyuan~Cheng, Hong, Li, Xiaoming~Wang, Xu, and Chi]{mixed_negative_sampling_goog_www20}
Yang, J., Yi, X., Zhiyuan~Cheng, D., Hong, L., Li, Y., Xiaoming~Wang, S., Xu, T., and Chi, E.~H.
\newblock Mixed negative sampling for learning two-tower neural networks in recommendations.
\newblock In \emph{Companion Proceedings of the Web Conference 2020}, WWW '20, pp.\  441–447, 2020.
\newblock ISBN 9781450370240.

\bibitem[Zhai et~al.(2011)Zhai, Lou, and Gehrke]{hdss_sigmod11}
Zhai, J., Lou, Y., and Gehrke, J.
\newblock Atlas: A probabilistic algorithm for high dimensional similarity search.
\newblock In \emph{Proceedings of the 2011 ACM SIGMOD International Conference on Management of Data}, SIGMOD '11, pp.\  997–1008, 2011.
\newblock ISBN 9781450306614.

\bibitem[Zhai et~al.(2023{\natexlab{a}})Zhai, Gong, Wang, Sun, Yan, Li, and Liu]{meta23ndp}
Zhai, J., Gong, Z., Wang, Y., Sun, X., Yan, Z., Li, F., and Liu, X.
\newblock Revisiting neural retrieval on accelerators.
\newblock In \emph{Proceedings of the 29th ACM SIGKDD Conference on Knowledge Discovery and Data Mining}, KDD '23, pp.\  5520–5531, New York, NY, USA, 2023{\natexlab{a}}. Association for Computing Machinery.
\newblock ISBN 9798400701030.
\newblock \doi{10.1145/3580305.3599897}.
\newblock URL \url{https://doi.org/10.1145/3580305.3599897}.

\bibitem[Zhai et~al.(2023{\natexlab{b}})Zhai, Jiang, Wang, Jia, Zhang, Chen, Liu, and Zhu]{bytetransformer2022}
Zhai, Y., Jiang, C., Wang, L., Jia, X., Zhang, S., Chen, Z., Liu, X., and Zhu, Y.
\newblock Bytetransformer: A high-performance transformer boosted for variable-length inputs.
\newblock In \emph{2023 IEEE International Parallel and Distributed Processing Symposium (IPDPS)}, pp.\  344--355, Los Alamitos, CA, USA, may 2023{\natexlab{b}}. IEEE Computer Society.
\newblock \doi{10.1109/IPDPS54959.2023.00042}.
\newblock URL \url{https://doi.ieeecomputersociety.org/10.1109/IPDPS54959.2023.00042}.

\bibitem[Zhang et~al.(2022)Zhang, Luo, Liu, Li, Chen, Zhang, Wei, Hao, Tsang, Wang, Liu, Li, Badr, Park, Yang, Mudigere, and Wen]{zhang2022dhen}
Zhang, B., Luo, L., Liu, X., Li, J., Chen, Z., Zhang, W., Wei, X., Hao, Y., Tsang, M., Wang, W., Liu, Y., Li, H., Badr, Y., Park, J., Yang, J., Mudigere, D., and Wen, E.
\newblock Dhen: A deep and hierarchical ensemble network for large-scale click-through rate prediction, 2022.

\bibitem[Zhao et~al.(2018)Zhao, Xia, Zhang, Ding, Yin, and Tang]{drl_pagewise_recsys18}
Zhao, X., Xia, L., Zhang, L., Ding, Z., Yin, D., and Tang, J.
\newblock Deep reinforcement learning for page-wise recommendations.
\newblock In \emph{Proceedings of the 12th ACM Conference on Recommender Systems}, RecSys '18, pp.\  95–103, New York, NY, USA, 2018. Association for Computing Machinery.
\newblock ISBN 9781450359016.
\newblock \doi{10.1145/3240323.3240374}.
\newblock URL \url{https://doi.org/10.1145/3240323.3240374}.

\bibitem[Zhao et~al.(2023)Zhao, Yang, Wang, Liu, Shi, Hu, Zhang, and Yang]{zhao2023ucranking}
Zhao, Z., Yang, Y., Wang, W., Liu, C., Shi, Y., Hu, W., Zhang, H., and Yang, S.
\newblock Breaking the curse of quality saturation with user-centric ranking, 2023.

\bibitem[Zhou et~al.(2018)Zhou, Zhu, Song, Fan, Zhu, Ma, Yan, Jin, Li, and Gai]{din_baba_kdd18}
Zhou, G., Zhu, X., Song, C., Fan, Y., Zhu, H., Ma, X., Yan, Y., Jin, J., Li, H., and Gai, K.
\newblock Deep interest network for click-through rate prediction.
\newblock KDD '18, 2018.

\bibitem[Zhou et~al.(2020)Zhou, Wang, Zhao, Zhu, Wang, Zhang, Wang, and Wen]{s3rec_cikm20}
Zhou, K., Wang, H., Zhao, W.~X., Zhu, Y., Wang, S., Zhang, F., Wang, Z., and Wen, J.-R.
\newblock S3-rec: Self-supervised learning for sequential recommendation with mutual information maximization.
\newblock In \emph{Proceedings of the 29th ACM International Conference on Information \& Knowledge Management}, CIKM '20, pp.\  1893–1902, New York, NY, USA, 2020. Association for Computing Machinery.
\newblock ISBN 9781450368599.
\newblock \doi{10.1145/3340531.3411954}.
\newblock URL \url{https://doi.org/10.1145/3340531.3411954}.

\bibitem[Zhuo et~al.(2020)Zhuo, Xu, Dai, Zhu, Li, Xu, and Gai]{otm_icml20}
Zhuo, J., Xu, Z., Dai, W., Zhu, H., Li, H., Xu, J., and Gai, K.
\newblock Learning optimal tree models under beam search.
\newblock In \emph{Proceedings of the 37th International Conference on Machine Learning}, ICML'20. JMLR.org, 2020.

\end{thebibliography}
